\documentclass[lettersize,journal]{IEEEtran}
\usepackage{amsmath,amsfonts}
\usepackage{algorithmic}
\usepackage{algorithm}
\usepackage{array}
\usepackage[caption=false,font=normalsize,labelfont=sf,textfont=sf]{subfig}
\usepackage{textcomp}
\usepackage{stfloats}
\usepackage{url}
\usepackage{verbatim}
\usepackage{graphicx}
\usepackage{cite}
\usepackage{todonotes}
\usepackage{threeparttable}
\usepackage{arydshln}
\usepackage{booktabs}
\usepackage{subcaption}
\usepackage{placeins}
\usepackage{multicol}

\usepackage{hyperref}

\usepackage{amsmath}
\usepackage{amssymb}
\usepackage{bm}
\usepackage{multirow}

\usepackage{booktabs}
\usepackage{xspace}
\usepackage{diagbox}

\usepackage{url}

\usepackage{graphicx}
\usepackage{lipsum}
\usepackage{microtype}
\usepackage{nicefrac}
\usepackage{verbatim}
\usepackage{bbding}
\usepackage{enumitem}

\usepackage[capitalize]{cleveref}
\crefname{section}{Sec.}{Secs.}
\Crefname{section}{Section}{Sections}
\Crefname{table}{Table}{Tables}
\crefname{table}{Tab.}{Tabs.}

\usepackage{xcolor}

\makeatletter
\DeclareRobustCommand\onedot{\futurelet\@let@token\@onedot}
\def\@onedot{\ifx\@let@token.\else.\null\fi\xspace}
\def\eg{\emph{e.g}\onedot} 
\def\ie{\emph{i.e}\onedot}

\renewcommand{\paragraph}{%
	\@startsection{paragraph}{4}{\z@}%
	{0.1em \@plus 0.5ex \@minus 0.2ex}{-1em}%
	{\normalsize\bf}%
}
\newcommand\rurl[1]{%
  \href{https://#1}{\nolinkurl{#1}}%
}
\makeatother

\newcommand{\bx}{\bm{x}}
\newcommand{\by}{\bm{y}}

\begin{document}

\title{Revisit the Imbalance Optimization \\in Multi-task Learning: An Experimental Analysis}

\author{Yihang Guo*, Tianyuan Yu*, Liang Bai, Yanming Guo, Yirun Ruan, William Li\dag, Weishi Zheng
\thanks{Yihang Guo*, Tianyuan Yu*, Liang Bai, Yanming Guo, Yirun Ruan are with the Laboratory for Big Data and Decision, National University of Defense Technology, Changsha 410073, China (e-mail: guoyihang0714@163.com; ty.yu@nudt.edu.cn; bailiang\_nudt@163.com; guoyanming@nudt.edu.cn; ruanyirun@163.com).}
\thanks{William Li\dag{} is with the School of Computer Science and Engineering,
Sun Yat-sen University, Guangzhou 510275, China (e-mail: william.li1992@outlook.com).}
\thanks{Weishi Zheng is with the School of Computer Science and Engineering,
Sun Yat-sen University, Guangzhou 510275, China, and also with Peng Cheng
Laboratory, Shenzhen 518005, China (e-mail: wszheng@ieee.org).}
\thanks{* These authors contribute equally to this work.}
\thanks{\dag{} Corresponding Author}}

\markboth{Journal of \LaTeX\ Class Files,~Vol.~14, No.~8, August~2021}%
{Shell \MakeLowercase{\textit{et al.}}: A Sample Article Using IEEEtran.cls for IEEE Journals}

\IEEEpubid{0000--0000/00\$00.00~\copyright~2021 IEEE}


\maketitle


\begin{abstract}
Multi-task learning (MTL) aims to build general-purpose vision systems by training a single network to perform multiple tasks jointly. While promising, its potential is often hindered by "unbalanced optimization", where task interference leads to subpar performance compared to single-task models. To facilitate research in MTL, this paper presents a systematic experimental analysis to dissect the factors contributing to this persistent problem. Our investigation confirms that the performance of existing optimization methods varies inconsistently across datasets, and advanced architectures still rely on costly grid-searched loss weights. Furthermore, we show that while powerful Vision Foundation Models (VFMs) provide strong initialization, they do not inherently resolve the optimization imbalance, and merely increasing data quantity offers limited benefits. A crucial finding emerges from our analysis: a strong correlation exists between the optimization imbalance and the norm of task-specific gradients. We demonstrate that this insight is directly applicable, showing that a straightforward strategy of scaling task losses according to their gradient norms can achieve performance comparable to that of an extensive and computationally expensive grid search. Our comprehensive analysis suggests that understanding and controlling gradient dynamics is a more direct path to stable MTL than developing increasingly complex methods.

\end{abstract}


\section{Introduction}

\begin{figure*}[!t]
  \centering
  \includegraphics[width=0.98\textwidth]{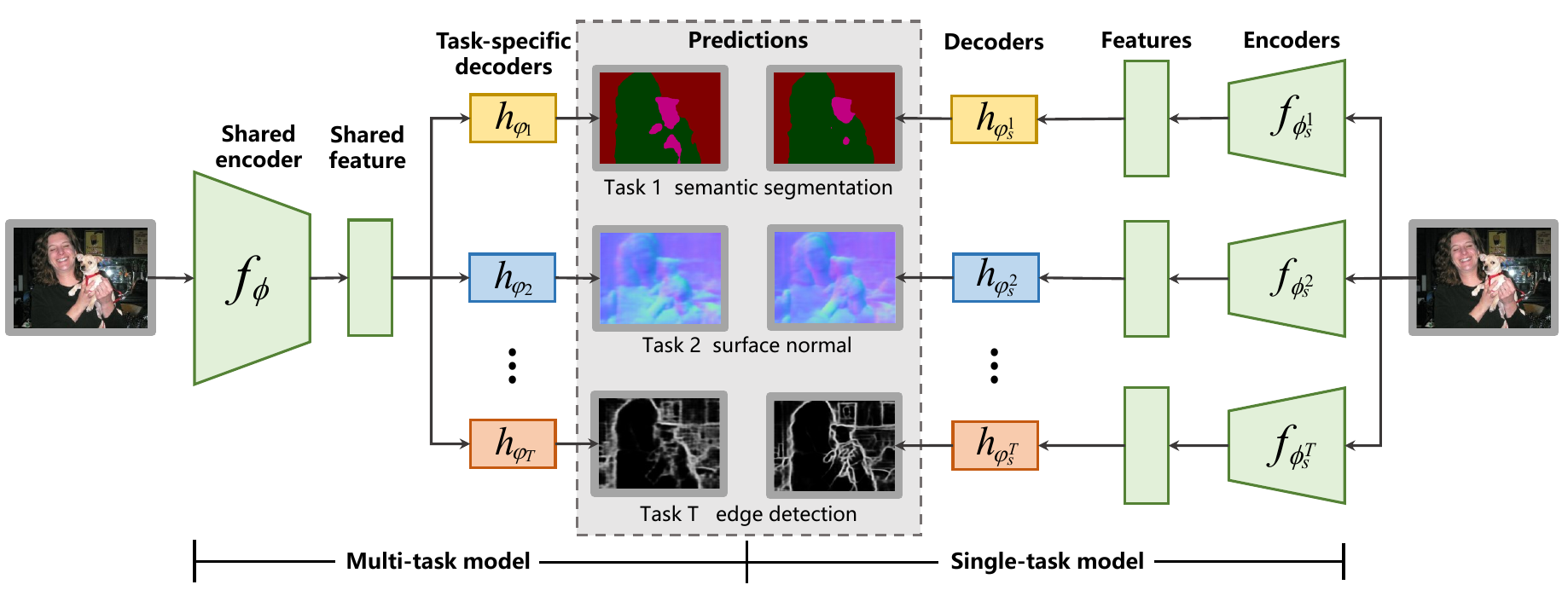}
  \caption{\textbf{Multi-task Learning (MTL) vs Single-task Learning (STL).} MTL aims to learn a single network that shares the encoder across tasks (left) while STL learns one network per task (right). Simply optimizing MTL involves in jointly minimizing multiple tasks losses is challenging and can lead to a severe imbalance problem, \ie, some tasks dominate the training and interfere the rest, resulting in worse performance on some times compared to learning a network per task (Single-task Learning).
      }
  \label{fig:diagram}
\end{figure*}

\IEEEPARstart{I}{n} the current era of artificial intelligence (AI), the frontier is rapidly advancing from narrow, specialized systems towards general-purpose foundation models. At the heart of this evolution lies the paradigm of Multi-Task Learning (MTL)~\cite{mtl_1997}, where a single, unified model is trained to solve a multitude of problems concurrently. This approach is fundamental to the capabilities of today's most influential technologies, from Large Language Models (LLMs) that are fine-tuned to simultaneously master translation, summarization, and reasoning, to the sophisticated perception systems in autonomous vehicles that must concurrently detect objects, parse scenes, and predict motion. By learning shared representations, MTL facilitates knowledge transfer across these diverse tasks, promising greater computational efficiency and superior generalization (\cref{fig:diagram}(left)) compared to training isolated, single-task networks (\cref{fig:diagram}(right)). However, despite its central role and promise, a significant and persistent challenge in MTL is the phenomenon of "imbalanced optimization", where interference among tasks during joint training often leads to degraded performance on certain tasks (\eg, results in \cref{tab:mto_performance}). Given the increasing reliance on MTL for building state-of-the-art AI, overcoming this imbalance is a critical step toward unlocking the full potential of these powerful, generalist models.


\begin{figure}[!t]
  \centering
  \includegraphics[width=0.50\textwidth]{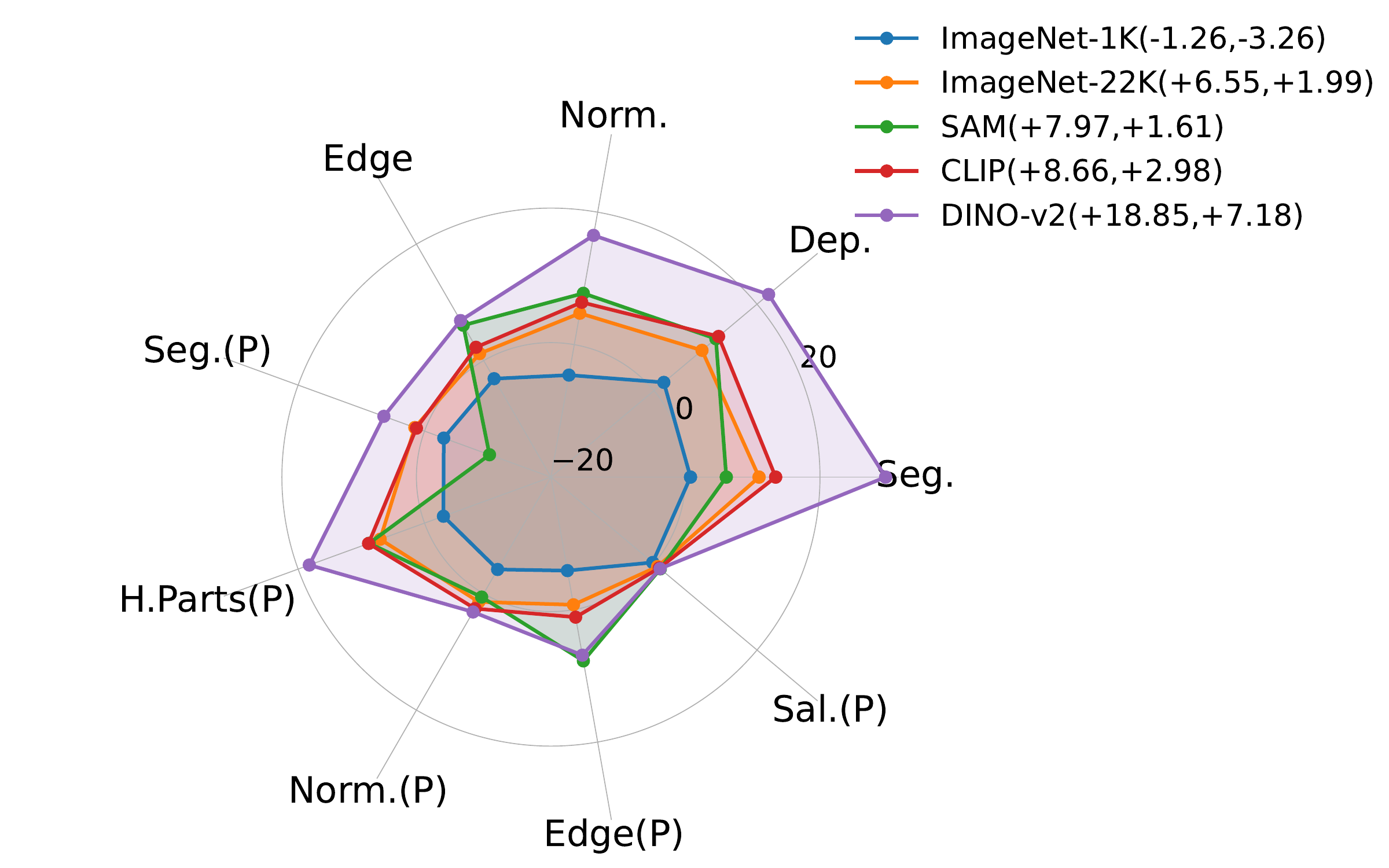}
    \caption{{Multi-task performance w.r.t. single-task learning performance of VFMs on NYUD-v2 and PASCAL-Context (P) datasets. VFMs' gain $\Delta_m$ (average improvement for all tasks) on two datasets are shown in the legend, respectively. We can see that SAM has the best performance on the edge detection task, but the results on the semantic segmentation task are much worse than other VFMs. }}
      \label{fig: intro_imbalance}
\end{figure}

The challenge of imbalanced optimization has led to a proliferation of specialized methods, largely divided into two categories.
The first focuses on the training dynamics themselves, with methods that either dynamically weight task losses to ensure a balanced contribution from each~\cite{uncert,mgda} or directly manipulate gradients to mitigate destructive interference during backpropagation~\cite{gradnorm,pcgrad,cagrad}. The second line of work tackles the problem at the architectural level, designing sophisticated networks to control information flow and reduce task-specific conflicts. \IEEEpubidadjcol Techniques in this vein range from learning adaptive network branches~\cite{bruggemann2020automated,Vandenhende_branched_2019,adashare} and employing task-specific attention layers~\cite{Cross-stitch,mtan}, to explicitly modeling cross-task relations across multiple scales~\cite{mti-net,atrc} and leveraging transformers for long-range correlations~\cite{invpt,Ye_Xu_2023}. Despite the progress driven by these specialized optimizers and architectures, task imbalance remains a significant and often unpredictable challenge. The full extent to which these methods, alongside other fundamental factors like model initialization and data properties, influence the optimization landscape remains unclear, motivating a more systematic and foundational analysis.

In response, we conduct this systematic analysis, aiming to build a more unified understanding of the task imbalance problem. While prior analytical studies have offered valuable insights~\cite{royer2023scalarization, MTO_helpful?, elich2023challenging,ieee_access}, they have predominantly focused on the efficacy of different optimization strategies. This narrow scope overlooks the complex interplay between the optimizer, the network architecture, the choice of pre-trained model, and the characteristics of the data itself. In this paper, we conduct a systematic and comprehensive analysis of the imbalanced optimization problem, specifically within the demanding context of dense computer vision. Our study extends beyond optimizers to empirically evaluate the distinct roles and interactions of: (1) network architectures, (2) the influence of large-scale Vision Foundation Models (VFMs) as initializations, and (3) the impact of data quality and quantity.

Our analysis begins by empirically establishing the prevalence and characteristics of the task imbalance issue across standard benchmarks. We first confirm that while vanilla MTL often underperforms single-task learning (STL), the imbalance can be effectively mitigated through a costly grid search for optimal loss weights. We then evaluate the two dominant lines of research intended to solve this problem automatically: multi-task optimization (MTO) methods and specialized architectures. Our findings reveal that the performance of existing MTO methods is often brittle and inconsistent across datasets. Similarly, while advanced architectures generally outperform STL, their success is frequently contingent on the same costly grid search for loss weights, indicating they do not fundamentally resolve the optimization problem.

This prompts a deeper investigation into more foundational factors. We analyze the influence of powerful Vision Foundation Models (VFMs) like CLIP~\cite{clip}, SAM~\cite{sam}, and DINO~\cite{DINO,dinov2}, and find that while they provide excellent initialization, they do not prevent the optimization imbalance from emerging (See \cref{fig: intro_imbalance}). We also show that simply increasing data quantity and quality has a limited effect. This analysis culminates in a key insight: the optimization imbalance is strongly correlated not with the angle of gradient conflict, but with the norm of task-specific gradients. We demonstrate that this insight is directly actionable, as a simple strategy of scaling losses to balance these gradient norms achieves performance comparable to the exhaustive grid search, pointing to a more direct and efficient path toward stable MTL.

In summary, our key contributions include: 

\begin{enumerate}
    \item We provide a comprehensive analysis of state-of-the-art optimization and architectural methods for multi-task dense prediction learning, revealing that their performance is often inconsistent across datasets and they frequently still rely on computationally expensive grid-searched loss weights to achieve good results.
    \item We analyze the effects of foundational factors beyond specific methods, including initialization with modern Vision Foundation Models (VFMs), data quality, and gradient properties. Our analysis shows that VFM choice and data characteristics have a limited effect, while identifying a strong correlation between the norm of task gradients and the imbalance problem.
    \item Motivated by our analysis, we demonstrate that the key to mitigating imbalance lies in controlling gradient norms. We show that a simple strategy of scaling losses according to their gradient norm consistently achieves performance comparable to an exhaustive grid search, offering an efficient and principled alternative.

\end{enumerate}

\begin{figure*}[!ht]
  \centering
  \includegraphics[width=0.88\textwidth]{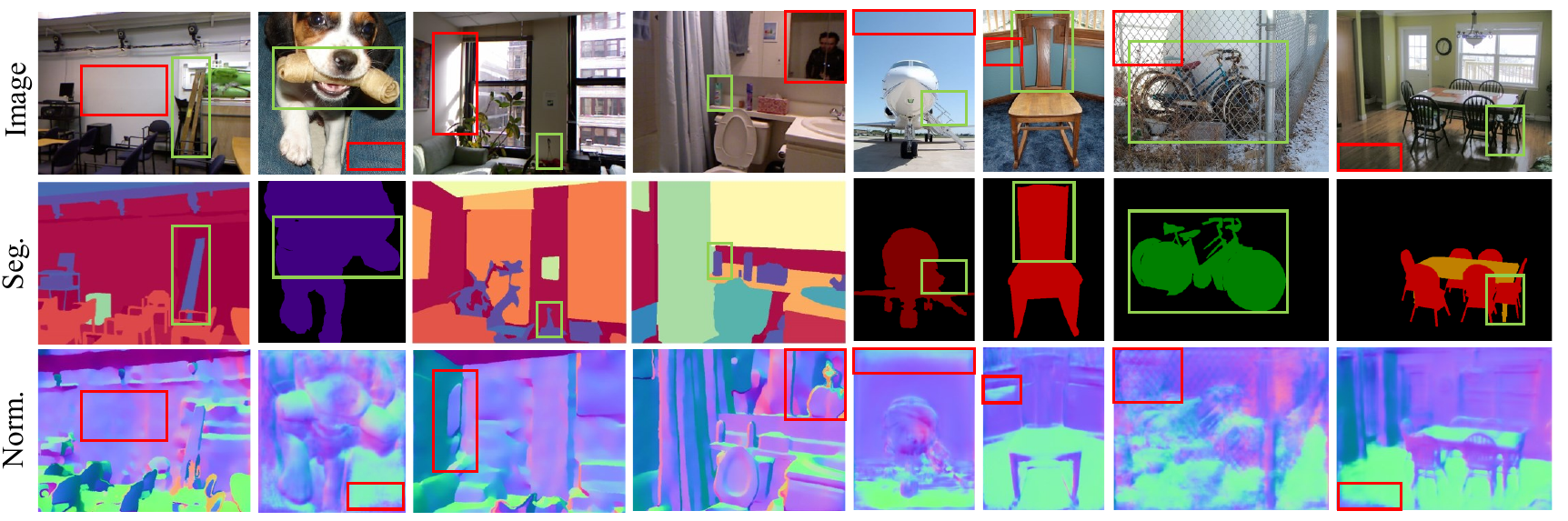}
  \vspace{-0.1cm}
  \caption{Few examples from NYUD-v2 and PASCAL-Context. Regions in segmentation (green) and surface normal (red) are noisy, causing the inconsistent in labels between tasks.}
  \label{fig:intro_noisy_example}
\end{figure*}

The rest of paper is organized as follows. Section II reviews related work. Section III establishes the problem definitions and categorizes potential solutions. Section IV provides experimental details while Section V details our comprehensive analysis of state-of-the-art MTL methods and foundational factors, and based on our findings, demonstrates the efficacy of a simple weight updating strategy. Section VI presents our conclusions.

\section{Related Work}

\begin{table*}[t]
    \centering
    \renewcommand{\arraystretch}{1.5}
    \caption{Recent studies in analyzing the imbalanced issue in multi-task learning.}
    \label{tab:example_table}
  \scalebox{0.98}{
  \begin{tabular}{ccl}
    \toprule
        Reference & Topics$^1$ &Limitations\\
    \midrule
        \cite{ieee_access}& MTO, Data&The sample size used to analyze the impact of the data on MTL was too small.\\
        \cite{Which_Tasks_Learned_Together?}& Data&Using only simple data sampling and viewing low data quantity as detrimental to MTL.\\
        \cite{MTO_helpful?}& MTO&The analysis of MTL optimization is on a fair scale but not deep enough.\\
        \cite{kurin2023defense}& MTO, Data&The effect of label noise is not considered in the corrupted data.\\
        \cite{elich2023challenging}& MTO&A simpler analysis to measure gradient conflict.\\
        \cite{royer2023scalarization}& MTO&Insufficient definition and analysis of gradient conflicts.\\
          \midrule
     Ours& Arch., MTO, Data and Pre-training&A comprehensive experiments on optimization, architectures,  generated data, and model initialization.\\
     \bottomrule
  \end{tabular}}
        \begin{tablenotes}
        \scriptsize
        \item 1. Research Topics in Multi-task Learning. Arch.: Architecture, MTO: Multi-task optimization.
         \end{tablenotes} 
        \label{tab:example_table}
\end{table*}

\subsection{Multi-Task Balanced Optimization} 
Our paper is related to the multi-task optimization (MTO) methods and these methods can be broadly divided into three groups.
\subsubsection*{Loss weighting}
The first group focuses on dynamically adjusting the weight of each loss term in MTL.
Uncertainty Weighting (UW)\cite{uncert} defines a learnable noise parameter to represent the uncertainty of a task and assigns lower weights to tasks with high (noisy) uncertainty. Dynamic Weight Average (DWA)\cite{mtan}) adjusts the loss weights by directly calculating the relative rate of decline in losses. Further, FAMO\cite{liu2023famo} is based on historical information about task losses and a dynamic weight adjustment mechanism that ensures that task losses decline at approximately equal rates. Dynamic Task Prioritization (DTP)\cite{guo2018dynamic} dynamically adjusts the weights according to task importance or learning stage. AUTO-lamada\cite{liu2022auto_lambda} uses the performance on a hold-out set to adjust the loss weights.
In addition, some studies consider the gradient information of the loss function under the shared parameters to adjust the task weights. GradNorm\cite{gradnorm} considers the learning rate of the task and the gradient norm of the shared layer. Multiple Gradient Descent Algorithm (MGDA) treats MTL as a multi-objective optimization problem to find a Pareto optimal solution. NashMTL\cite{nashmtl} optimizes the weights of each task through a bargaining game to ensure that the gradient contribution of each task is relatively balanced in the joint update direction.

\subsubsection*{Gradient update} Alternatively, the second group addresses the problem by modifying the gradients of loss functions w.r.t. the shared parameters to alleviate the conflicts among tasks. PCGrad\cite{pcgrad} projects the gradient of one task to the normal vector of the gradient of another task to eliminate gradient conflicts. GradDrop\cite{graddrop} randomly drops these gradient conflicts during training. CAGrad\cite{cagrad} considers minimizing the loss for all tasks and constrains the update of the aggregated gradient to a specific interval range. RotoGrad\cite{javaloy2022rotograd} rotates the shared feature space by introducing a task-specific rotation matrix. FairGrad\cite{fairgrad} transforms MTL optimization into a utility maximization problem by introducing the $\alpha$-fairness framework. However, such methods consume a lot of computational resources to capture the gradient information, and how to reduce the computational cost is one of the important challenges in this direction.

\subsubsection*{Distillation approaches and analysis on MTO methods} Recently, some researchers have argued that randomization~\cite{rlw} or unitary loss weights~\cite{MTO_helpful?,kurin2023defense} can be sufficient, suggesting that MTO is not always necessary to improve MTL performance. 
Apart from them, some researchers have consider knowledge distillation for distilling the knowledge in task-specific models into the multi-task network such that the multi-task network can achieve similar results with task-specific networks.
URL\cite{li2023Universal} aligns multiple feature representations of a task-specific network into a common feature space through knowledge distillation. With the proposed VFMs on a large scale\cite{clip,sam,dinov2}, it has been shown that high performance can be achieved for multiple downstream tasks by distilling multiple pre-trained vision base models into a unified model\cite{RADIO}. Swiss Army Knife (SAK)\cite{sak} preserves and adapts the unique representation bias of each VFM through the Teacher-Specific Adapter Path (TSAP) and uses Mixture of Representations Routers to generate customized features for each downstream task.


In this work, we aim at exploring the imbalanced problem in MTL, conducting an extensive evaluation of MTO methods. Our observations indicate that while uniform loss weighting can lead to imbalanced performance, extensive loss weight searching and certain MTO methods can improve performance, albeit with some exhibiting inconsistent performance across benchmarks. 
Furthermore, unlike prior studies~\cite{royer2023scalarization, MTO_helpful?, elich2023challenging,ieee_access} where only multi-task optimization methods are analysized, we analyze the impact of other factors, including optimization, architecture, and data, on the imbalanced optimization problem in MTL. A detailed comparison with existing studies is depicted in \cref{tab:example_table}.

\subsection{Multi-Task Architecture} 

Another popular solution to address the imbalance optimization is to design architectures that effectively share relevant information while retaining task-specific elements~\cite{uberNet,latent,Stochastic_Filter_Groups,Vandenhende_branched_2019,multinet_2016,pad_net,pap-net,atrc,network_search_2018,jtrl_2018,Task_Routing}. And they can be broadly divided into two categories.

\subsubsection*{Encoder layer} Encoder-based methods focus on feature extraction and sharing strategies for backbone networks. ~\cite{Vandenhende_branched_2019,adashare} treats different tasks by designing network branching mechanisms. ~\cite{Cross-stitch} learns inter-task weights through cross-task attention mechanisms. ~\cite{mtan} Dynamically selects the most useful features for the task at hand through a soft-attention mechanism. ~\cite{bruggemann2020automated,Guo_Lee_Ulbricht_2020}controls feature propagation between tasks via a gating module. The expert hybrid module ~\cite{NEURIPS2022_b653f34d,Ye_Xu_2023,Chen_2023_CVPR,jiang2024mlore} performs task-specific prediction by combining the outputs of different experts. ~\cite{taskprompter2023,Ye_Xu_2023} dynamically adjusts the model's response to different tasks via learnable cue vectors.

\subsubsection*{Decoder layer} Decoder-based methods focus on how to extract the most useful information for each task from a shared feature representation. PAD-Net\cite{pad_net} combines task features from different heads through distillation units for final prediction. PAP-Net\cite{pap-net} uses task features from different heads to compute pixel affinity matrices for each task. The affinity matrices are adaptively combined and diffused back into the task feature space to propagate cross-task related information across images. MTI-Net\cite{mti-net} utilizes multi-scale feature maps to convey complementary information between tasks. The synergy between tasks is enhanced by cross-task modules. ATRC\cite{atrc} achieves comprehensive and effective task interactions through Bridge Features. InvPT\cite{invpt} uses features from different stages for decoding layer interactions through long range spatial correlation achieved by Transformer.

In this work, we focus on analyzing the effects of the architecture on imbalanced optimization in MTL. Our analysis encompasses both convolutional neural network (CNN) and vision transformer (ViT) based methods, aiming to provide insights into how architectural choices can influence the task imbalance issue.

\subsection{Other Multi-Task Learning Research} 
Multi-task learning has been widely applied in many fields, including recommendation systems~\cite{Recommender}, computer vision~\cite{simon_survey}, natural language processing~\cite{nlp}, multi-modal multi-task learning~\cite{munikoti2024generalist}(also known as generalist). The MTL field encompasses a range of research interests such as task grouping~\cite{Which_Tasks_Learned_Together?,task_group_2021efficiently}, cross-task relations~\cite{3D-Aware_li2023multitask,zamir2020robust,taskonomy}, many tasks~\cite{cappart2023combinatorial,kurin2020can,Stochastic_Filter_Groups}, multi-domain~\cite{bilen2017universal,rebuffi2018efficient,mensink2013distance}, continual~\cite{rosenfeld2018incremental}, curriculum~\cite{narvekar2020curriculum}, meta-learning~\cite{hospedales2021meta,wang2022semi}, semi/partially-supervised MTL~\cite{li2022Learning}, and multi-model distillation~\cite{mizrahi20244m}. 

In contrast to these diverse areas, our work takes an orthogonal direction by focusing on the analysis of factors leading to task imbalance and we hope that our findings can be helpful for these fields.

\section{Imbalanced Problem in Multi-task Learning and Existing Methods}\label{sec:statement}
In this section, we briefly review both single-task and multi-task learning, discussing the imbalanced problem and existing optimization and architecture methods. We focus on MTL in the supervised learning setting.

Let $\mathcal{D}$ be a training set composed of $N$ RGB training images and their respective labels. We assume that the training images are sampled from a single distribution (or dataset), and each training image $\bx$ is associated with labels $Y$ for $T$ tasks, where $Y=\{\by_1, \ldots, \by_T\}$. 

\subsection{Single-task Learning (STL)}

In single-task learning, we aim to learn a set of models $\{y_t(x) = h_{\psi_t} \circ f_{\phi_t}(\bx)\}$, each for a task, for addressing $T$ tasks in the $\mathcal{D}$ datasets. Each model $y_t$ consists of a feature encoder $f_{\phi_t}: \mathbb{R}^{3 \times H \times W} \rightarrow \mathbb{R}^{C \times H' \times W'}$ parameterized by $\phi_t$ to map an image to a feature map, which is decoded to the prediction for task $t$ by a decoder $h_{\psi_t}: \mathbb{R}^{C \times H' \times W'} \rightarrow \mathbb{R}^{O_t \times H_t \times W_t}$. We learn a model for each task independently:
\begin{equation}\label{eq:stl}
\min _{\phi_t,\psi_t} \frac{1}{N} \sum_{\bx, \by_t} \mathcal{L}_t\left(\hat{y}_t\left(\bx\right), \by_t\right),
\end{equation}
where $\mathcal{L}_t$ is the loss function for task $t$.

\subsection{Multi-task learning (MTL)}
The vanilla multi-task learning model shares a common feature encoder $f_{\phi}$ across all tasks, followed by task-specific decoders $h_{\psi_t}$ for generating predictions for different tasks: $y_t(\bx) = h_{\psi_t} \circ f_{\phi}(\bx)$. The MTL models is learned by jointly minimizing multiple tasks' losses:
\begin{equation}\label{1}
\min _{\phi,\left.\left\{\psi_t\right\}\right|_{t=1} ^T} \frac{1}{N} \sum_{\bx, \by_1, \cdots, \by_T}\lambda_t \mathcal{L}_t\left(\hat{y}^t\left(\bx\right), \by_t\right),
\end{equation}
where $\lambda_t$ is the loss weight for task $t$ and $\lambda_t=1$ for all tasks is the vanilla MTL. 

\subsection{Imbalanced problem in MTL}

Jointly minimizing multiple tasks losses to optimize a single network (Multi-task learning network) as in \cref{1} often leads to subpar performance compared to their single-task learning counterpart, i.e., each network is trained by optimizing its task loss as in \cref{eq:stl}. We observe and discuss this phenomenon in \cref{sec:analysis}. In in multi-task learning, we call this challenge of achieve similar or better results than the single-task networks the imbalanced problem~\cite{uncert,li2023Universal}. As multi-task learning aims at learn a single model jointly tackling multiple tasks while also achieving good performance in all tasks compared with single-task learning, researchers in this field have explore strategies for addressing the imbalanced problem in MTL. 


\subsection{Existing Methods for Addressing the Imbalance Problem}
\subsubsection*{Multi-task optimization} To alleviate the negative effect, prior research efforts such as ~\cite{uncert,mgda} have attempted to dynamically estimate the task-specific loss weights $\lambda_t$ to achieve a more balanced optimization process during MTL. 

To optimize the MTL model which consists of a shared feature encoder $f_{\phi}$ followed by multiple task-specific decoders, the parameter update w.r.t. the shared encoder's parameter $\phi$ typically involves the aggregation of gradients from multiple tasks losses:
\begin{equation}\label{eq:gradient}
    \phi \leftarrow \phi - \eta \sum_{t=1}^T \lambda_t \nabla_{\phi}\mathcal{L}_t,
\end{equation}
where $\nabla_{\phi}\mathcal{L}_t$ is the derivative of the loss $\mathcal{L}_t$ of task $t$ with respect to the shared parameters $\phi$. Simply setting $\lambda_t$ in \cref{eq:gradient} to be the same for any tasks during the optimization often lead to the imbalanced optimization and conflicts among gradients. To address the issue, some prior works, such as CAGrad~\cite{cagrad}, propose gradient update strategies and UW~\cite{uncert}, design a mechanism that dynamically estimates loss weights $\lambda_t$. We denote these methods as $G(\cdot)$: 
\begin{equation}\label{eq:G}
\phi \leftarrow \phi - \eta \sum_{t=1}^T G(\nabla_{\phi}\mathcal{L}_t).
\end{equation}

\subsubsection*{Multi-task achitectures}
Apart from optimization strategies in MTL, some prior work has also focuses on designing architectures for better sharing information across tasks for mitigating the imbalance problem. 
These architectures can be formulated as $y_t=h_{\psi_t, \beta_t} \circ f_{\phi, \alpha_t}$, where $\alpha_t$ represents task-specific designs in the feature encoder and $\beta_t$ is the designed module to better accumulate information across tasks during the decoding phase. Here, $\alpha_t$ can be the attention modules in MTAN~\cite{mtan} that is attached to the shared encoder to extract task-specific features or the feature propagation modules in MTI-Net~\cite{mti-net} for modeling cross-task interaction within multi-scale features. In MTI-Net~\cite{mti-net}, $\beta_t$ is the prediction distilling modules to estimate the final task-specific prediction via aggregating information from multiple tasks' predictions. 

\subsubsection*{Other strategies}
Alternatively, with the success in recent vision foundation models (VFMs), the VFMs are viewed as a good multi-task learner or initialization for addressing the imbalance problem. 

Despite these methods, we show that the imbalance problem persists and the effects of optimization, architectures and VFMs on MTL remain unclear due to the lack of through and systemaic analysis. To this end, in the next section, we perform systematically analysis on these factors for addressing the imbalance problem in MTL.




\section{Experimental Details}\label{sec:experiments}





\subsection{Experimental setups}

\subsubsection*{Datasets} 
We conduct the experiments on two widely-used multi-task datasets: NYUD-v2\cite{nyud} and PASCAL-Context\cite{pascal}, and a synthetic dataset Replica\cite{replica} from Omnidata~\cite{omnidata}
The details for these datasets are described below.

\begin{enumerate}
    \item \textbf{NYUD-v2} contains RGB-D indoor-scene images.
    We follow the standard settings where 795 images are used for training and 654 images are for testing. We follow the prior work~\cite{invpt,3D-Aware_li2023multitask} and use labels for four tasks: semantic segmentation (Seg.), monocular depth estimation (Dep.), surface normal estimation (Norm.), and edge detection (Edge).
    
    \item \textbf{PASCAL-Context}, the benchmark subset we adopt from the PASCAL dataset, is widely used for dense prediction tasks. This dataset contains 4,998 training samples and 5,105 testing samples, annotated for five tasks of semantic segmentation, human parts segmentation (H. Parts), surface normal estimation, saliency detection (Sal.) and edge detection. Among them, the surface normal and saliency labels were distilled by previous work\cite{Maninis_2019_CVPR}.
    \item \textbf{Repilca} is a 3D dataset for reconstruction containing 18 high fidelity scenes. We use the data provided by Omnidata~\cite{omnidata}. 
    This dataset consists of 56783 images for training, 23725 images for validating, and 13889 images for testing. We consider three common tasks in vision: semantic segmentation, monocular depth estimation, and surface normal estimation.
\end{enumerate}

\subsubsection*{Evaluation Metrics} We follow the prior work~\cite{simon_survey,invpt,atrc} and use the same evaluation metrics. Specifically, we use mean Intersection over Union (mIoU) for segmentation problems, Maximal F-measure (maxF) for saliency detection, optimal dataset F-measure (odsF) for edge detection, root mean square error (RMSE) for monocular depth estimation, and mean error (mErr) for surface normal estimation. 
To quantify the average performance gain of the MTL methods, we also use the multi-task performance evaluation metrics as in prior work~\cite{simon_survey}. $\Delta_m$ defined as the average per-task performance improvement relative to a single-task baseline:
\begin{equation}\label{eq:delta}
\Delta_{m} = \frac{1}{T}\sum_{i=1}^TM_{i},
\end{equation}
where $T$ is the number of tasks. $M_{i}=(-1)^{l_i}(M_{m,i}-M_{b,i})/M_{b,i}$ represents the performance metric for task $i$, and $l_i$ is a binary indicator, equal to 1 if a lower metric value signifies better performance for task $i$, and 0 otherwise. 

\subsubsection*{Model Backbone} We use three commonly used backbones in our experiments as prior state-of-the-art MTL methods. 
Specifically, we use ResNet~\cite{resnet}, HRNet~\cite{hrnet} and ViT~\cite{vit} respectively. Following the prior work~\cite{simon_survey}, we add dilated convolutions to ResNet to extract high resolution features for dense prediction tasks. We also use three popular MTL architectures which are MTAN~\cite{mtan}, MTI-Net~\cite{mti-net} and InvPT~\cite{invpt} based on ResNet, HRNet and ViT respectively. For MTI-Net, we further construct a more close single-task learning baseline for fair comparisons with architecture based methods as deep supervision (estimating prediction at earlier layers and applying supervision), feature propagation module across multiple scales, prediction distillation in architecture-based MTL methods (MTI-Net) can also benefit single-task learning network. For the multi-task prediction distillation module, we modified it to self-distillation in STL.




\subsubsection*{Training details} We conduct all experiments on Pytorch with NVIDIA RTX4090 GPU. For ResNet, we train all models on NYUD-v2 and PASCAL-Context for 100 and 60 epochs, respectively. For HRNet and ViT, we train all models on NYUD-v2 and PASCAL-Context for 40k iterations. On the Replica, we use InvPT~\cite{invpt} as the base model and follow all the settings in InvPT~\cite{invpt} for all variants that use ViT as backbone. If not specifically mentioned, all experiments are performed using pretrained ImageNet weights. Please refer to the supplementary for more details.



\section{Experimental Analysis}\label{sec:analysis}

\begin{table*}[t]
    \centering
    \renewcommand{\arraystretch}{1.2}
    \caption{Performance of multi-task learning relative to single-task learning on NYUD-v2 and PASCAL-Context with ResNet-18.}
    \label{tab:mto_performance}
    \scalebox{1}{
    \begin{tabular}{cc|ccccc|cccccc}
        \toprule
        \multicolumn{1}{c}{\multirow{3}{*}{Backbone}} & \multirow{3}{*}{Method}& \multicolumn{5}{c|}{NYUD-v2}& \multicolumn{6}{c}{PASCAL-Context}\\
        \cmidrule{3-13}&& Seg.& Dep.& Norm.& Edge&\multirow{2}{*}{$\Delta_m$\%↑}& Seg.& H. Parts& Norm.& Sal.& Edge& \multirow{2}{*}{$\Delta_m$\%↑}\\
         && mIoU↑ & RMSE↓ & mErr↓ & odsF↑  && mIoU↑& mIoU↑ & mErr↓ & maxF↑ & odsF↑ & \\
        \midrule
        \multirow{15}{*}{ResNet-18}&Single-task& 39.38& 0.6111& \textbf{21.94}& 72.40&+0.00& 66.59 & \textbf{61.29} & \textbf{13.67} & 84.77& \textbf{66.90} & \textbf{+0.00}\\
        & Uniform  & 39.70& 0.6030& 23.37& 67.10&-2.93& 66.52& 60.02& 14.99& 83.25& 62.50&-4.04\\
 \cdashline{2-13}
        
        &Grid Search  & 39.87& 0.5999& 22.74& 72.40&-0.15& 65.10& 59.28& 14.39& 83.91& 65.50& -2.64\\
        &UW\cite{uncert}  & 39.90& 0.6084& 23.86& 67.70&-3.37& 66.25 & 60.01 & 15.60 & 83.93 & 64.30 & -4.15\\
        &GradNorm\cite{gradnorm}  & 39.31& \textbf{0.5939}& 22.95& 68.20&-1.95& 64.30 & 59.31 & 14.73 & 84.29 & 65.60 & -3.38 \\
        &DWA\cite{mtan}  & 39.66& 0.5961& 23.28& 67.40&-2.46& \textbf{66.75} & 60.14 & 14.92 & 83.23 & 62.60 & -3.80 \\
        &MGDA\cite{mgda}  & 38.89& 0.6071& 22.61& 66.80&-2.85& 61.48 & 54.41 & 13.95 & 82.13 & 58.90 & -7.19 \\
        &RLW\cite{rlw}  & 39.29& 0.6068& 23.57& 66.40&-3.82& 64.83 & 59.04 & 15.38 & 82.56 & 60.40 & -6.23 \\
        &FAMO\cite{liu2023famo}  & 39.02& 0.6110& 23.34& 67.20&-3.62& 66.39& 59.99& 14.97& 83.04& 62.30& -4.17 \\
        &PCGrad\cite{pcgrad}  & 39.12& 0.6001& 23.22& 67.40&-2.90& 66.17 & 60.46 & 14.82 & 83.32 & 63.60 & -3.41 \\
        &CAGrad\cite{cagrad}  & 40.22& 0.6052& 23.03& 72.80&-0.34& 65.95 & 59.98 & 14.98 & 83.24 & 66.60 & -2.98 \\
 & GradDrop\cite{graddrop}  & 39.38& 0.6020& 23.28& 67.00&-3.02& 66.69 & 59.93 & 14.95 & 83.02 & 61.10 &-4.42 \\
 & URL\cite{li2023Universal}& \textbf{40.17}& 0.5940& 22.70& 71.00&-0.16& 66.61& 60.82& 14.19& \textbf{84.96}& 65.30&-1.32\\
 & NashMTL\cite{nashmtl}& 39.59& 0.5979& 22.37& \textbf{72.90}&+0.35& 66.37& 60.60& 14.65& 84.23& 66.50&-2.40 \\
 & FairGrad\cite{fairgrad}& 39.91& 0.5953& 22.37& 72.70&\textbf{+0.59}& 64.69& 59.82& 14.71& 83.43& 66.10&-3.12 \\
 \midrule
 \multirow{15}{*}{MTAN}& Single-task  & 39.42& 0.6054& \textbf{21.75}& 73.40&+0.82& 67.52& \textbf{62.83}& \textbf{13.56}& \textbf{84.77}& 65.20&\textbf{+0.45}\\
& Uniform  & \textbf{41.15}& 0.6072& 22.60& 72.90&+0.69& 67.84& 61.23& 14.58& 83.44& 65.60&-1.67\\

 \cdashline{2-13}
 
 & Grid Search  & 40.27& 0.6020& 22.02& 73.70&+1.29& 65.46& 60.47& 14.14& 84.43& 67.40&-1.22\\
 & UW\cite{uncert}  & 40.29& 0.6055& 22.89& 72.60&-0.21& \textbf{67.87}& 61.30& 14.83& 83.90& 67.00&-1.48\\
 & GradNorm\cite{gradnorm}  & 41.00& 0.6091& 22.40& 72.80&+0.72& 65.84& 60.12& 14.63& 84.23& 67.30&-2.01\\
 & DWA\cite{mtan}  & 40.88& 0.6051& 22.62& 72.80&+0.56& 67.70& 61.46& 14.53& 83.25& 65.80&-1.55\\
 & MGDA\cite{mgda}  & 40.05& 0.6025& 22.28& 72.80&+0.53& 62.66& 55.69& 13.91& 81.99& 61.10&-5.74\\
 & RLW\cite{rlw}  & 40.74& 0.6036& 22.79& 72.70&+0.30& 65.59& 60.21& 15.08& 83.04& 63.90&-4.02\\
 & FAMO\cite{liu2023famo}  & 40.24& 0.6032& 22.78& 72.50&-0.05& 67.46& 61.07& 14.58& 83.56& 65.50&-1.87\\
 & PCGrad\cite{pcgrad}  & 40.37& 0.6007& 22.59& 72.60&+0.38& 66.97& 61.47& 14.51& 83.68& 65.70&-1.66\\
 & CAGrad\cite{cagrad}  & 40.97& 0.6077& 22.63& \textbf{73.90}&+1.01& 66.69& 60.92& 14.90& 83.39& \textbf{68.30}&-1.79\\
 & GradDrop\cite{graddrop}  & 40.96& 0.6051& 22.64& 72.70&+0.56& 66.79& 60.79& 14.78& 83.28& 64.50&-2.79\\
  & URL\cite{li2023Universal}& 40.18& \textbf{0.5912}& 21.97& 73.30&\textbf{+1.59}& 67.51& 61.34& 14.59& 83.55& 65.60&-1.72\\
   & NashMTL\cite{nashmtl}& 40.80& 0.6007& 22.17& 73.80&+1.54& 66.49& 61.05& 14.33& 84.49& 68.20&-0.74\\
    & FairGrad\cite{fairgrad}& 40.68& 0.6023& 22.16& 73.10&+1.17& 64.69& 59.82& 14.71& 83.43& 67.60&-1.61\\
        \bottomrule
    \end{tabular}
    }
\end{table*}

\subsection{How Pervasive is the Imbalance Problem in Vanilla MTL?
}
We first conduct experiments of training a single network per task, \ie, single task learning \emph{STL} and also the vanilla multi-task learning with uniform (\emph{Uniform}) and grid searched (\emph{Grid Search}) loss weights where the encoder is shared across tasks and is followed by task-specific decoders. The results on NYUDv2 and PASCAL-Context are reported in \cref{tab:mto_performance}.
From the results we can see that simply jointly minimizing multiple task loss in MTL (Uniform) lead to \emph{imbalance results}, \eg, Uniform obtains good performance in some tasks (\eg depth in NYUDv2) while it performs much worse on the rest (\eg surface normal and edge in NYUDv2), compared with single-task learning networks. And we also observe that this imbalanced problem can be mitigated by a better loss weighting approach, \ie Grid Search which we extensively search the optimal loss weight and Grid Search enables the vanilla MTL methods to get close or better results than single-task learning networks (\eg results in NYUDv2). 

This suggests that it is possible that MTL can obtains better or comparable results than STL if the imbalanced problem is addressed.
However, intensively searching a set of optimal loss weights requires large cost and is timeconsuming and can be suboptimal as we are unable to intensively search the optimal loss weights dynamically.


\subsection{How Effective are Existing Optimization (MTO) Methods at Mitigating Imbalance?}

To mitigate the imbalance problem in MTL, researchers in MTL has designed several optimization strategies to dynamically estimating better loss weights, gradient updates or aligning MTL with STLs: 

\begin{enumerate}
    \item \textbf{loss weighting} methods aim to dynamically estimating a loss weight per task to balanced the optimization of multiple loss functions according to the uncertainty of the each task prediction (UW~\cite{uncert}), gradients' norms~\cite{gradnorm}, loss decreasing speed (DWA~\cite{mtan}, FAMO~\cite{liu2023famo}), random weights (RLW~\cite{rlw}), pareto solution (MGDA~\cite{mgda}) and Nash Bargaining Solution (NashMTL\cite{nashmtl}).
    \item \textbf{gradient updating} strategies aim at addressing the imbalance problem by avoiding the opposite directions of gradients, including gradient surgery (PCGrad~\cite{pcgrad}), conflict-averse gradient descent (CAGrad~\cite{cagrad}), gradient sign dropout (Graddrop~\cite{graddrop}), and fairness aware gradient descent (FairGrad\cite{fairgrad}). 
    \item \textbf{distillation} approaches instead mitigate the imbalance problem by aligning the features of MTL and a set of STL models such that the MTL achieves similar or better results than STL models, \eg URL\cite{li2023Universal}. 
\end{enumerate}



In light of recent literature~\cite{simon_survey,MTO_helpful?,kurin2023defense} that questions the benefits of MTO methods, we conduct the experiments that compare most existing MTO methods with both \emph{Uniform} and \emph{Grid Search}. 
Results based on the ResNet-18 backbone are reported in  \cref{tab:mto_performance}. 

First, we can see that several MTOs (e.g., NashMTL, FairGrad) demonstrate significant superiority over the Uniform, with FairGrad achieving superior performance (+0.59\% on NYUD-v2). This indicates that multi-task optimization can effectively mitigate task imbalance compared to uniform or random loss weighting. Compared to Grid Search, only a subset of methods (e.g., CAGrad) show superior performance in specific tasks (e.g., Edge). Apart from NashMTL, few approaches consistently outperform Grid Search across all scenarios. These observations suggest that Grid Search remains a competitive strategy, and identifying optimal task-specific loss weights persists as a challenging yet practically significant objective in MTL research.

Second, we observe that some methods perform worse in most scenarios (e.g., MGDA and RLW), where the performance of MGDA is as low as -7.19\% in PASCAL-Context, with even larger imbalances across tasks (e.g., Seg. and Norm.). MGDA enforces a constrained relationship between gradient direction and update direction for each task, and this strict Pareto constraint may be more limited in multi-task optimization. RLW balances the losses among tasks by randomly sampling the loss weights, and this randomness may not accurately reflect the true importance among tasks, leading to over-amplification or under-amplification of the losses in some tasks. RLW deviates from the current mainstream assumptions for MTL imbalance optimization (e.g., gradient conflict or gradient alignment).

Third, we observe the additive effect of a strong backbone on multi-task optimization. The performance of MTOs generally improves when using a stronger backbone (ResNet-50 on NYUD-v2, see \cref{tab:mto res50 on nyud}) or a specific network architecture designed for MTL (MTAN). This suggests that stronger backbone provides better shared feature representations that better capture inter-task correlations and provide a better basis for MTOs.


Finally, though it is still very challenging for MTL method to outperform single-task network in all tasks, we find that designing an optimization that estimating optimal loss weights or gradient updates can effectively improves the performance without increasing the inference cost. Among MTOs, they are inconsistent across datasets and grid searched loss weights are most stable and obtains comparable results. This suggests that it is essential to design more powerful optimization techniques to address the imbalance problem in the future.

\begin{table}
    \centering
    \renewcommand{\arraystretch}{1.2}
    \caption{Performance of multi-task learning relative to single-task learning on NYUD-v2 with ResNet-50.}
    \label{tab:mto res50 on nyud}
    \scalebox{1}{
    \begin{tabular}{c|ccccc}
        \toprule
         \multirow{3}{*}{Method}& \multicolumn{5}{c}{NYUD-v2}\\
        \cmidrule{2-6}& Seg.& Dep.& Norm.& Edge& \multirow{2}{*}{$\Delta_m$\%↑}\\
         & mIoU↑ & RMSE↓ & mErr↓ & odsF↑  & \\
        \midrule
        Single-task& 42.94& 0.5905& \textbf{20.54}& 73.40& +0.00\\
  Uniform  & 43.70& 0.5733& 21.71& 73.50&  -0.22\\

   \hdashline
   
        Grid Search  & 43.98& 0.5735& 21.47& \textbf{74.60}&  +0.60\\
        UW\cite{uncert}  & 44.00& 0.5754& 22.10& 73.80&  -0.76\\
        GradNorm\cite{gradnorm}  & 44.55& 0.5801& 21.49& 73.70&  +0.58\\
        DWA\cite{mtan}  & 44.00& 0.5764& 21.59& 73.80&  +0.32\\
        MGDA\cite{mgda}  & 43.33& 0.5776& 21.29& 73.70&  +0.21\\
        RLW\cite{rlw}  & 43.69& 0.5816& 21.74& 73.40&  -0.64\\
        FAMO\cite{liu2023famo}  & 43.66& 0.5754& 21.76& 73.80&  -0.57\\
        PCGrad\cite{pcgrad}  & 44.28& 0.5713& 21.62& 73.90&  +0.67\\
        CAGrad\cite{cagrad}  & 43.84& 0.5789& 21.70& \textbf{74.60}&  +0.02\\
        GradDrop\cite{graddrop}  & 43.53& 0.5804& 21.71& 73.70&  -0.55\\
        URL\cite{li2023Universal}& \textbf{44.98}& \textbf{0.5598}& 20.84& 74.20&  \textbf{+1.91}\\
        NashMTL\cite{nashmtl}& 43.93& 0.5741& 21.17& \textbf{74.60}&  +0.62\\
        FairGrad\cite{fairgrad}& 43.97& 0.5694& 21.16& 74.40&  +0.76\\
        \bottomrule
    \end{tabular}
    }
\end{table}

\begin{figure*}[!ht]
  \centering
  \includegraphics[width=1\textwidth]{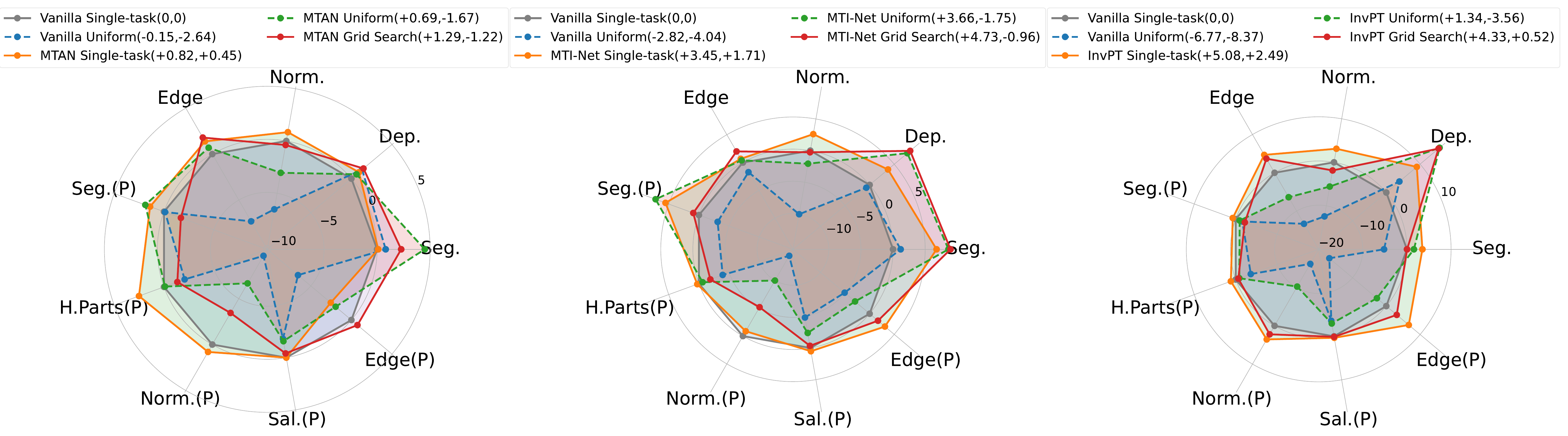}
  \caption{Radar charts of MTL performance for MTAN, MTI-Net, InvPT, and their base model ResNet-18, HRNet-18, and ViT-T/16 on NYUD-v2 and PASCAL-Context (P) dataset. The coordinate values represent the improvement of each task relative to a Vanilla Single-task, i.e.$M_{i}$. MTL architectures Gain $\Delta_m$ on two datasets are shown in the legend, respectively.}
  \label{archi}
\end{figure*}

\subsection{Do Specialized Architectures Fundamentally Solve the Imbalance Problem?}


Another group of solutions for imbalanced optimization in MTL involves introducing distinct shared modules for different tasks. So we also evaluate four widely recognized architecture-based MTL approaches, including two encoder-focused method, \ie vanilla MTL and MTAN~\cite{mtan}, and two decoder-focused methods, MTI-Net~\cite{mti-net} and InvPT~\cite{invpt}. Based on vanilla MTL, MTAN uses additional attention modules to extract task-specific information; MTI-Net instead design modules to model cross-task conrrelations across multi-scale features; InvPT also consider cross-task relations over multi-scale features while it extract long-range spatial correlations via transformer modules. 

The improvements of all methods w.r.t. the Vanilla-Single task are depicted in Figure~\ref{archi}. Our observations are as follows. 
First, Existing MTL architectures (green dashed polygon) significantly improves MTL performance over the Vanilla MTL with Uniform loss weights (blue dashed polygon). This indicate that the cross-tasks correlation designs in existing architecture methods can alleviate the imbalanced optimization problem.
Second, All the architecture-based methods with uniform loss weights outperform Vanilla Single-task on the NYUD-v2 dataset, but all fall short on the more challenging PASCAL-Context dataset (\eg, the MTL performance of MTAN with Uniform is positive for NYUD-v2(+0.69\%) and negative(-1.67\%) for PASCAL-Context);
At last, Compared with CNN based architecture (\eg, ResNet or HRNet), simply adopting ViT architecture for MTL by fully sharing the encoder across all tasks without any cross-task designs can lead to severer imbalanced optimization.

While architecture-based methods are often claimed superior performance over the vanilla MTL models without any designed modules, many compare their architectures against STL. This may not accurately reflect the effectiveness of their proposed solutions~\cite{mti-net} for addressing the imbalanced problem as some modules can be also beneficial for STL, e.g., extracting features from multi-scale features.
To this end, we also assess the performance of the STL with modules in the corresponding archiecture based MTL methods. For example, we implement the STL of MTI-Net, denoted as MTI-Net Single-task, by adapting and adding the Feature Propagation Module (FPM) and deep supervision proposed in MTI-Net into the vanilla single-task model. 


Comparing MTAN, InvPT, and MTI-Net with their respective new single-task model (\ie single-task network with adapted modules proposed in these methods, \eg, MTI-Net Single-task) on the NYUD-v2 and PASCAL-Context, our observations are as follows. 
(4) Existing MTL architectures (green dashed polygon) still obtains worse performance than their single-task learning models (orange solid polygon). Although MTI-Net Uniform (+3.66\%) outperforms MTI-Net Single-task (+3.45\%) on the NYUD-v2 dataset, there is still a significant gap on the PASCAL-Context dataset (-1.75\% vs +1.71\%). This indicating that the success of architecture-based models for MTL may lie in their unified approach to single task design. In addition. the gap between these MTL architectures (green vs orange polygon) is smaller than the one between the vanilla MTL and STL (blue vs gray polygon). For example, InvPT Uniform compared to InvPT Single-task is -3.74\% and -6.05\%, while the difference between vanilla MTL and STL is -6.77\% and -8.37\%. This indicates that architecture based MTL methods can still mitigating the imbalance problem while they still can not outperform single-task networks.
(5) Architecture based methods also rely on grid searched loss weights. 

In summary, the existing architecture can mitigate the imbalance optimization problem to some extent, but the effect is also limited. These findings suggest that both better architecture designs and optimization approaches can be complementary to help alleviate the unbalanced problem while the architecture based MTL methods still rely on the optimization approaches, e.g. existing methods rely on grid searched loss weights. Searching such loss weights or most existing MTO methods are still costly and unstable.
Besides, it is still challenging to reach the performance of single-task learning in all tasks and more powerful optimization techniques are encouraged. These findings and difficulties motivate us to throughly explore and study the effects of different factors on the unbalanced problem in MTL.

\subsection{Do Vision Foundation Models (VFMs) Prevent Optimization Imbalance?}
Vision Foundation Models are often considered as a good multi-task learner and are helpful for addressing the imbalance problem as they are trained using the same loss function instead of multiple distinct loss function that can cause the imbalance problem. Additionally, a recent work~\cite{pretrained_weight_conflict} also argue that the interference between tasks during optimization is caused by the initialization of the network. 

To this end, we also investigate the effects of pre-trained weights of the backbone. We conducted a comparative analysis using ViT-B models that were pre-trained on ImageNet-1K, ImageNet-21K, ImageNet-22K, CLIP, SAM, DINO, and DINO-v2 as initializations for the shared backbone for both vanilla MTL (Uniform and Grid Search) and STL on the NYUD-v2 and PASCAL-Context datasets. The results are depicted in Table~\ref{tab:pretrained weight}. 
We find that these VFMs, which have been shown to have strong generalized recognition capabilities, still face task imbalances, and no VFM can outperform its own single-task learning couterparts. However, it is still possible to achieve more balanced performance than uniform with a grid search strategy. Notably, we find significant representation bias in different VFMs towards different types of tasks as also observed in~\cite{sak} and they are learned in various ways and inherently has different biases.


Our conclusions show that even with the use of powerful VFMs as an initialization method, it is still difficult to overcome the task imbalance problem. Effectively utilizing the visual representation bias of different VFMs to further mitigate multi-task imbalance may be an interesting direction\cite{sak}. In addition, there can be imbalance problem in learning VFMs and exploring strategies for addressing imbalance problem can be also beneficial for VFMs.

\begin{table*}[!ht]
    \centering
    \renewcommand{\arraystretch}{1.2}
    \caption{Performance of multi-task learning relative to single-task learning on NYUD-v2 and PASCAL-Context with different VFMs.}
    \label{tab:pretrained weight}
    \scalebox{1}{
    \begin{tabular}{cc|ccccc|cccccc}
        \toprule
        \multicolumn{1}{c}{\multirow{3}{*}{Initialization}
} &\multirow{3}{*}{Method}& \multicolumn{5}{c|}{NYUD-v2}& \multicolumn{6}{c}{PASCAL-Context}\\
        
\cmidrule{3-13} 
&& Seg.& Dep.& Norm.& Edge&\multirow{2}{*}{$\Delta_m$\%↑}& Seg.& H. Parts& Norm.& Sal.& Edge& \multirow{2}{*}{$\Delta_m$\%↑}\\
          
&& mIoU↑ & RMSE↓ & mErr↓ & odsF↑  && mIoU↑& mIoU↑ & mErr↓ & maxF↑ & odsF↑ & \\
        \midrule
        \multirow{3}{*}{ImageNet-1K}&Single-task& 46.26& 0.6689& 20.59& 74.10&+0.00& 75.60& 62.21& 14.39& 84.40& 69.70& +0.00\\
        
&Uniform& 46.49& 0.6630& 22.25& 68.20&-3.67& 74.94& 60.87& 16.00& 83.44& 61.60&-5.40\\
 
&Grid Search& 46.61& 0.6561& 21.55& 71.80&-1.26& 73.30& 60.35& 14.99& 84.19& 65.60& -3.26\\

        \cdashline{1-13}\multirow{3}{*}{ImageNet-21K}&Single-task& 51.50& 0.6018& 19.17& 76.10&+7.74& 78.37& 68.93& 13.73& 85.28& 73.70& +5.17\\
        
&Uniform& 50.31& 0.6216& 20.16& 70.50&+3.26& 77.80& 67.71& 14.96& 84.52& 64.30& +0.04\\
        
&Grid Search& 51.03& 0.6084& 19.69& 74.90&+6.21& 76.39& 66.59& 14.21& 85.09& 69.10& +1.86\\

        \cdashline{1-13}\multirow{3}{*}{ImageNet-22K}&Single-task
& 51.40& 0.5937& 19.14& 76.30&+8.09& 79.08& 69.31& 13.73& 85.27& 73.70& +5.47\\
        
&Uniform
& 51.13& 0.6067& 20.09& 70.50&+4.36& 77.94& 67.71& 14.90& 84.39& 64.80& +0.27\\
        
&Grid Search
& 51.32& 0.6066& 19.61& 75.00&+6.55& 76.77& 66.57& 14.18& 84.99& 69.20& +1.99\\

   \cdashline{1-13}\multirow{3}{*}{CLIP}& Single-task
& 53.57& 0.5819& 18.88& 77.60& +10.46& 79.55& 70.54& 13.36& 85.19& 74.80& +6.81\\
 
& Uniform
& 52.76& 0.5658& 19.58& 71.50& +7.71& 78.05& 69.02& 14.48& 84.68& 66.10& +1.74\\
 
& Grid Search& 52.46& 0.5850& 19.26& 75.80& +8.66& 76.57& 67.72& 14.02& 85.27& 70.50& +2.98\\

 \cdashline{1-13}\multirow{3}{*}{SAM}& Single-task
& 47.01& 0.5666& 18.23& 79.10& +8.78& 71.72& 70.30& 13.60& 86.56& 77.60& +5.45\\
 
& Uniform
& 47.97& 0.5639& 19.74& 76.00& +6.52& 70.70& 68.06& 15.02& 84.23& 71.60& +0.22\\
 
& Grid Search
& 49.07& 0.5885& 18.99& 78.60& +7.97& 67.83& 67.66& 14.30& 85.44& 75.10& +1.61\\

 \cdashline{1-13}\multirow{3}{*}{DINO}& Single-task
& 46.29& 0.6104& 20.10& 75.80& +3.37& 74.53& 66.15& 13.91& 84.19& 72.80&+2.49\\
 
& Uniform
& 45.77& 0.5925& 20.81& 71.10& +1.31& 73.94& 64.72& 14.78& 84.46& 65.30&-1.43\\
 
& Grid Search
& 46.31& 0.6245& 20.54& 74.10& +1.75& 71.65& 63.16& 15.23& 83.89& 65.70&-1.30\\

 \cdashline{1-13}\multirow{3}{*}{DINO-v2}& Single-task
& 59.78& 0.4996& 16.41& 79.70& +20.60& 82.48& 76.46& 13.28& 86.17& 78.50& +10.89\\
      
& Uniform
& 59.29& 0.4971& 17.63& 75.60& +17.56& 81.46& 74.78& 14.57& 84.87& 70.20& +5.59\\
        &Grid Search& 60.03& 0.5201& 17.19& 79.20&+18.85& 80.46& 73.57& 13.98& 85.38& 74.50& +7.18\\
        
\bottomrule
    \end{tabular}
    }
\end{table*}

\subsection{Can Task Imbalance be Mitigated by Improving Data Quality or Quantity?}\label{sec: data}

In this section, we delve deeper into the role of data quality, and data quantity, to determine their impact on the imbalance problem in MTL. To achieve this, we adopt the synthetic dataset, \ie Replica, and we use the version from Omnidata~\cite{omnidata} with pixel-wise annotations for three tasks. The dataset has much large number of samples and much cleaner annotations than NYUD-v2 and PASCAL-Context datasets.

\begin{figure}[!ht]
    \centering
    \includegraphics[width=1\linewidth]{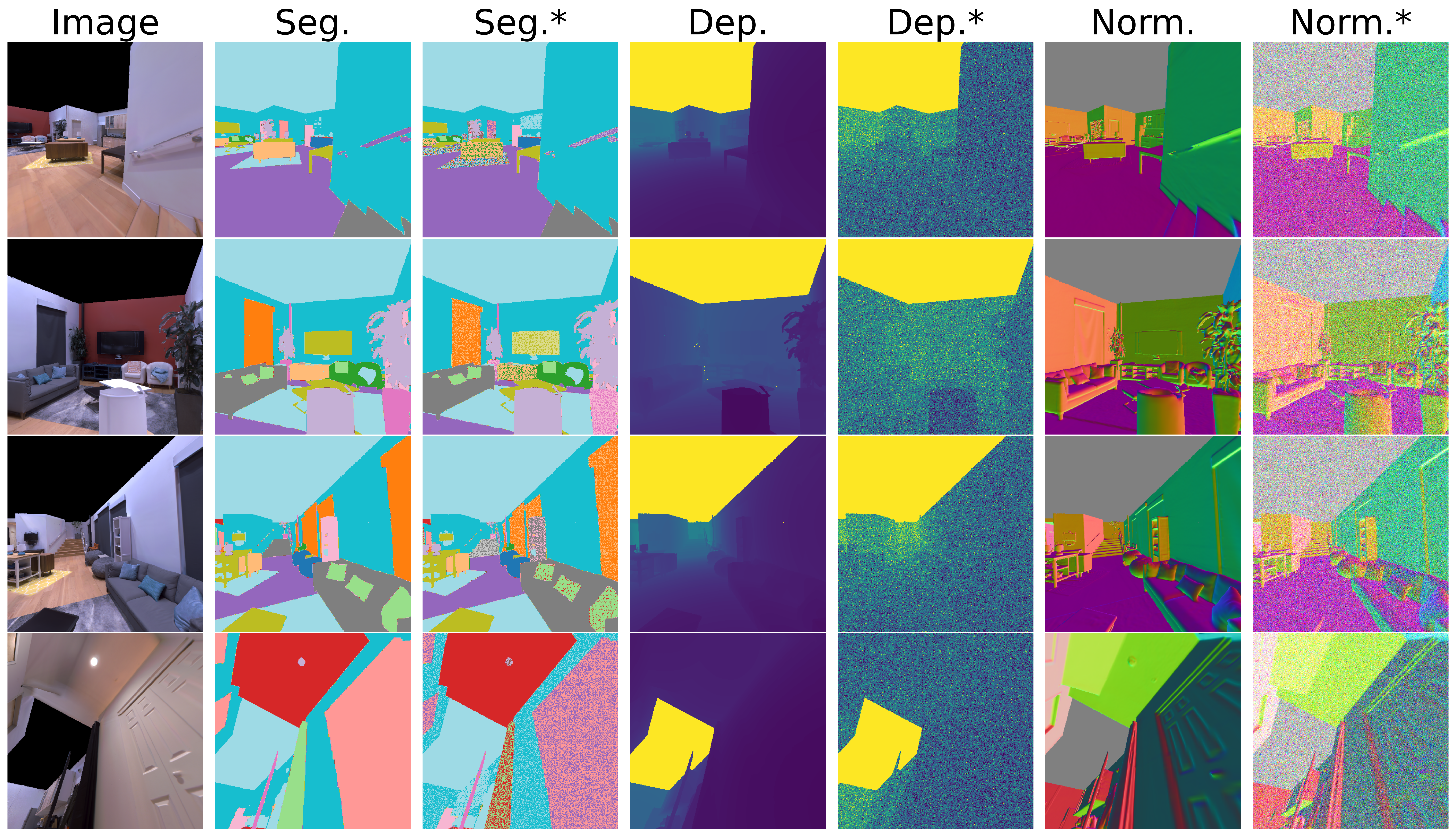}
    \caption{The first column of the original image. $^*$ is a different noise added for each task.}
    \label{fig:noisy_data}
\end{figure}

\begin{table*}[!ht]
    \centering
    \renewcommand{\arraystretch}{1.2}
  \caption{Results on the Replica dataset (3 tasks). We report results for adding label noise and using less data.}
  \label{replica_sample_ratio}
  \scalebox{1}{
  \begin{tabular}{ccc|cccc} 
              \toprule
               \multirow{2}{*}{Operation}&\multirow{2}{*}{Sample} &\multirow{2}{*}{Method}&Seg. &Dep. &Norm. &  \multirow{2}{*}{$\Delta_m$\%↑}\\
       &&& mIoU↑& RMSE↓& mErr↓&\\
    \midrule
            {\multirow{2}{*}{Clean}}&Full&Single-task&55.15& 0.4358& 21.09& +0.00\\
           &Full&Multi-task&53.14& 0.4455& 21.88& -3.21\\
   \cdashline{1-7}{\multirow{2}{*}{Noise}}& Full& Single-task& 48.21& 0.5055& 26.11&-17.46\\
 & Full& Multi-task& 43.27& 0.4971& 26.04&-19.69\\
 \cdashline{1-7}{\multirow{2}{*}{1/2 number of scans (per scene)}}& Less& Single-task& 53.86& 0.4285& 21.29&-0.54\\
     &  Less& Multi-task& 51.38& 0.4492& 21.27& -3.59\\
 \cdashline{1-7}{\multirow{2}{*}{1/4 number of scans (per scene)}}& Less& Single-task& 53.35& 0.4306& 21.28&-0.99\\
 & Less& Multi-task& 51.04& 0.4806& 21.39&-6.38\\
 \cdashline{1-7}{\multirow{2}{*}{12 scans (per scene)}}& Less& Single-task& 52.82& 0.4396& 21.86&-2.92\\
 & Less& Multi-task& 48.02& 0.5549& 22.05&-14.94\\
 \cdashline{1-7}{\multirow{2}{*}{1/2 number of scenes (all scans)}}& Less& Single-task& 53.35& 0.4299& 21.38&-1.10\\
 & Less& Multi-task& 51.14& 0.4889& 21.30&-6.82\\
 \cdashline{1-7}{\multirow{2}{*}{1/4 number of scenes (all scans)}}& Less& Single-task& 53.48& 0.4205& 21.31&-0.19\\
 & Less& Multi-task& 51.62& 0.4961& 21.27&-7.03\\
 \cdashline{1-7}{\multirow{2}{*}{1 scene (all scans)}}& Less& Single-task& 52.64& 0.4284& 21.60&-1.76\\
      &Less& Multi-task& 47.87& 0.5358& 22.09& -13.63\\ 
         \bottomrule
  \end{tabular}}\label{tab:data_analysis}
\end{table*}


\subsubsection*{Data Quality}
We firstly investigate the effect of the data quality by evaluating the STL and MTL models over the original data or the data with added noise. Here, we randomly add noise to every tasks annotations rather than the images. For semantic segmentation, for each sample, we randomly select five categories (or keep them all if there are less than five semantic labels) and change the pixels of the selected categories to the wrong category at a rate of 0.5. For depth and surface normal, we add gaussian noise. This would cause the inconsistent between different tasks' labels and will cause conflicts between tasks and can lead to the imbalanced optimization in MTL. Some examples of noisy task labels are shown in Figure \ref{fig:noisy_data}.


Results are reported in \cref{tab:data_analysis}. We can see that both the MTL and STL approaches achieve better performance when using clean data, while adding noise to different tasks leads to a larger negative migration of the model. This validates that noise leads to inconsistency in task annotations, which in turn leads to negative migration during the training of MTL model. However, there is no clear indication that noise exacerbates task imbalance, as training models for individual tasks independently that are also subject to noise reduces performance. 

\subsubsection*{Data Quantity} As Replica contains multiple scans from multiple scene, instead of randomly sampling different number of samples from the whole dataset, we randomly sample various number of scans per scene or different scenes. These all result in various samples for training. We train both MTL and STL from various number of samples and comparing with the STL and MTL model learned from all samples. From the results in \cref{tab:data_analysis}, we can see that STL and MTL models with less samples obtains worse performance. We also observe that the performance of MTL learned from less samples is much imbalanced than the one learned from all samples. This indicates that less samples can lead to severer imbalanced problem. However, training the model with more samples is not necessary for addressing the imbalance problem. 


\subsection{Further Analysis in Foundational Factors}

\begin{figure}[!t]
  \centering
  \includegraphics[width=0.48\textwidth]{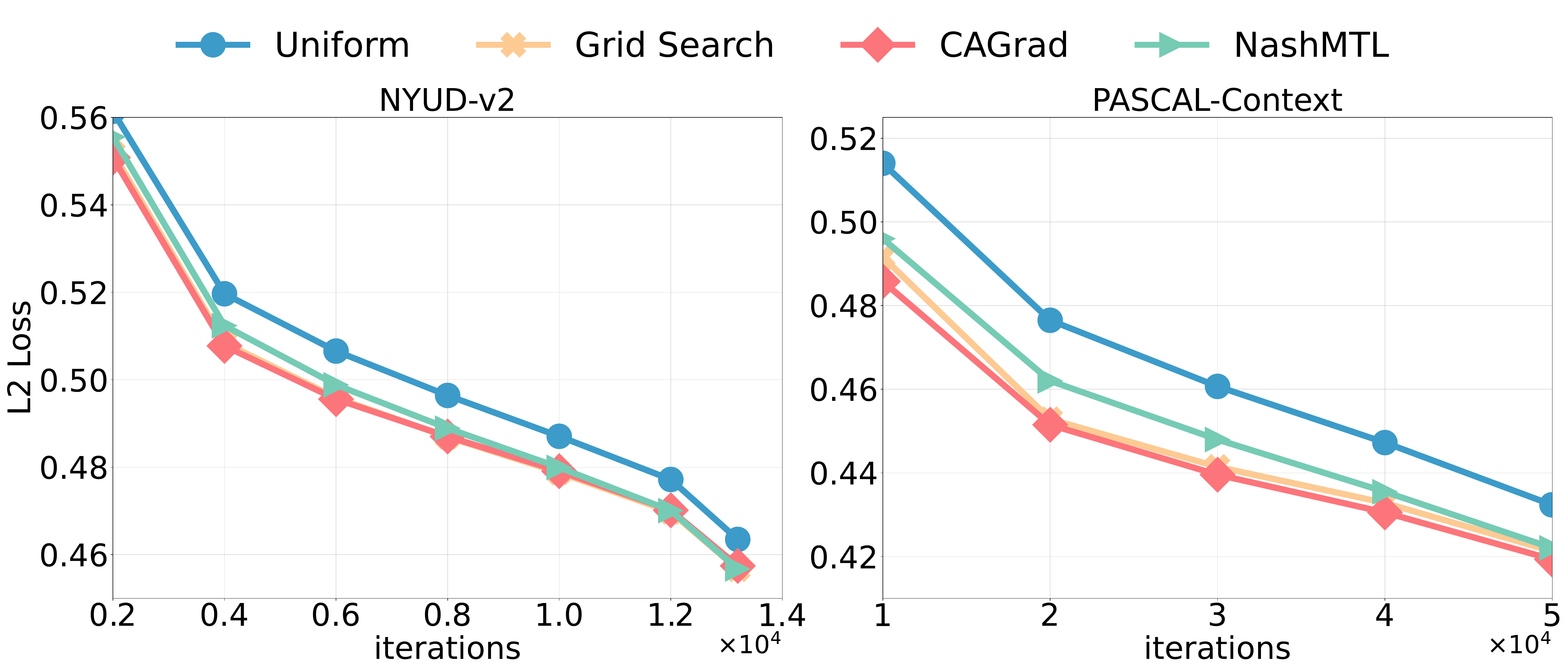}
  \caption{L2 loss of feature map between STL/MTL of different methods for each task on NYUD-v2 and PASCAL-Context.}
  \label{fig:feature_map}
\end{figure}

\textbf{1) Does Feature Similarity Correlate with MTL Performance?}
Inspired by \cite{li2023Universal}, we aim to study the differences in the feature map of different MTOs in STL/MTL, as shown in \cref{fig:feature_map}. Specifically, we first trained an STL model, and when training the MTL model, we updated the adaptor by calculating the feature differences between the STL and MTL (we use the L2 function to compute the distance as in \cite{li2023Universal}). We observed that Grid Search, CAGrad, and NashMTL obtained better performance and their feature differences were closer to the corresponding single-task, compared to Uniform (which obtained the worst $\Delta_m$).

\textbf{2) Does Feature Similarity Correlate with MTL Performance?}
Inspired by \cite{li2023Universal}, we aim to study the differences in the feature map of different MTOs in STL/MTL, as shown in \cref{fig:feature_map}. Specifically, we first trained an STL model, and when training the MTL model, we updated the adaptor by calculating the feature differences between the STL and MTL (we use the L2 function to compute the distance as in \cite{li2023Universal}). We observed that Grid Search, CAGrad, and NashMTL obtained better performance and their feature differences were closer to the corresponding single-task, compared to Uniform (which obtained the worst $\Delta_m$). 

This indicates that an effective multi-task model should strive to learn similar feature representations to individual single-task models. Moreover, using single-task models to guide the feature representation of multi-task models is proven to be effective.

\begin{figure*}[!ht]
  \centering
  \includegraphics[width=1\textwidth]{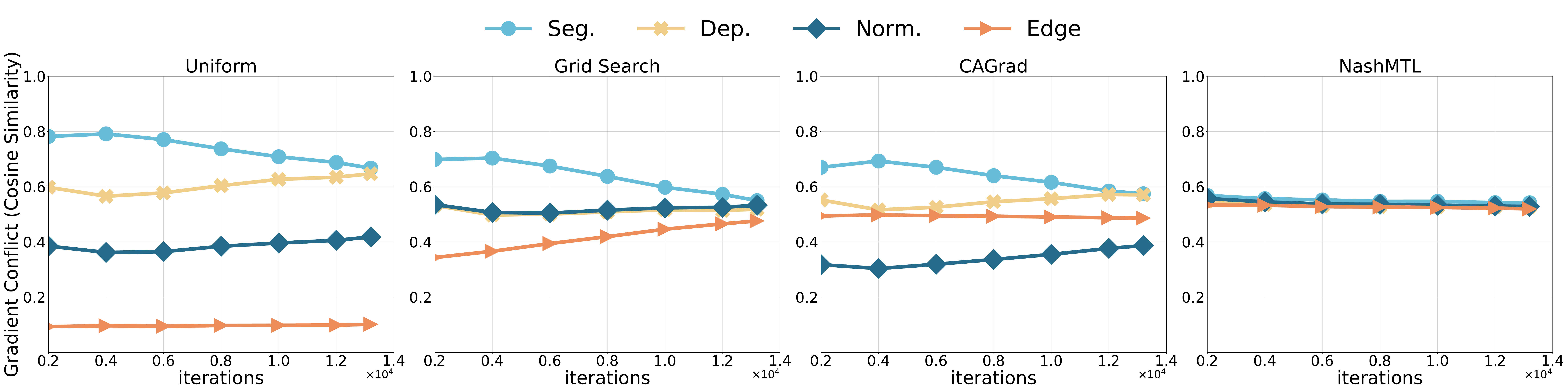}
  \caption{Gradient conflict between the gradient contributed by each individual task and the aggregated gradient for Uniform, Grid Search, CAGrad and NashMTL on NYUD-v2.}
  \label{fig:avg_grad_conflict_nyu}
\end{figure*}

\begin{figure*}[!ht]
  \centering
  \includegraphics[width=1\textwidth]{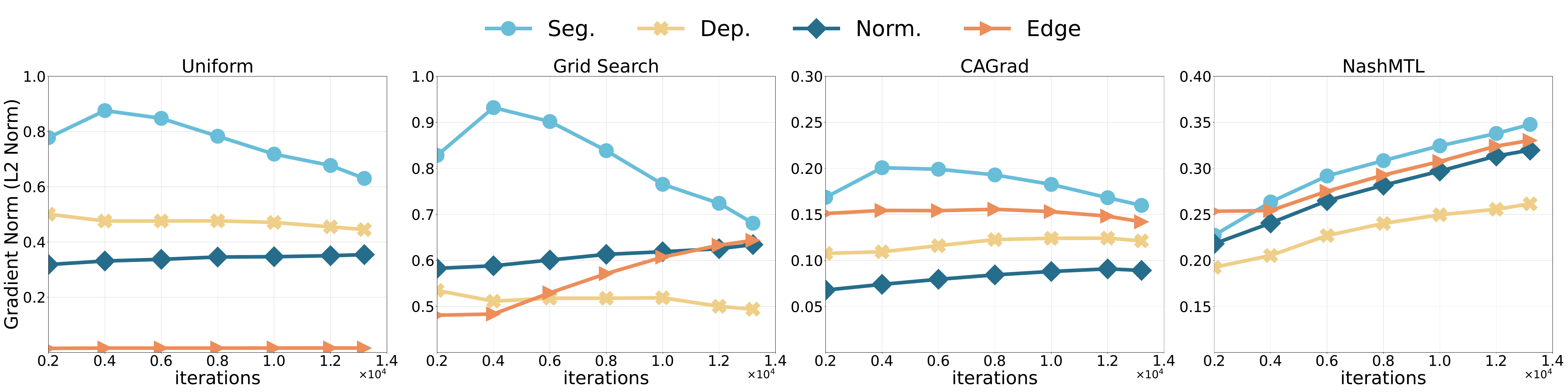}
  \caption{Gradient (w.r.t. the last layer of the encoder) norm for each task in Uniform, Grid Search, CAGrad, and NashMTL on NYUD-v2.}
  \label{fig:grad_norm_nyu}
\end{figure*}

As optimizing a multi-task learning methods involves manipulating multiple tasks gradients w.r.t. the shared parameters (\cref{eq:gradient}), the gradients conflicts may cause the imbalanced optimization problem in MTL~\cite{pcgrad}. To this end, we aim to visualize the gradient similarity and compare between the MTL methods with uniform, grid searched, CAGrad and NashMTL (Best performing MTO) searched loss weights.

\textbf{3) Is gradient similarity between tasks the key to understand imbalance?}
We first consider exploring the gradient similarity between tasks and we consider two ways for computing the gradient similarity. The first is similar to~\cite{pcgrad}, we use the cosine similarity to measure the conflict between different task gradients. The second is similar to~\cite{cagrad}, we use the dot product to measure the conflict between different task gradients. Based on our observations, the gradient similarity between tasks in MTL does not show a clear cue that mitigating conflict in gradients can mitigate the imbalance problem. We have analysed this in more detail in the appendix.

\textbf{4) Is gradient similarity between each task and the aggregated direction a Reliable Predictor of Imbalance?}
Alternatively, the relations between gradient of each task and the aggregated gradients can be related to the imbalance problem. To this end, we compute the cosine similarity between the gradient contributed by each individual task and the aggregated gradient, shown in \cref{fig:avg_grad_conflict_nyu}. 


For methods such as Grid Search, CAGrad, and NashMTL, which are considered more effective, the cosine similarity converges a similar value (\eg, 0.5) as training progresses. This suggests a balanced contribution of each task's gradient to the aggregated gradient. In contrast, the Uniform method, which does not perform as well, exhibits diverging curves, indicating that it is trapped in a state of task imbalance. 

This observation suggests that the key to overcoming task imbalance lies in the similarity between each task's gradient and the aggregated gradient. A more balanced contribution from each task, as indicated by a cosine similarity closer to 0.5, may be indicative of a more effective multi-task optimization strategy.

\textbf{5) Task-specific Gradient Norm is Crucial to Imbalance!}
Apart from the angle between gradients of different tasks, the scale (norm) of gradients were also considered as a important factor to imbalanced optimization problem~\cite{mgda}.
In \cref{fig:grad_norm_nyu}, we illustrate the change in gradient norm for each task during the training process on the NYUD-v2 dataset. 
We found that there seems to be a correlation between the norm of each task gradient and the MTL performance, \ie, the norm of the gradient for all tasks in the Grid Search, CAGrad and NashMTL (Best performing) methods also seem to converge to a similar value in NYUD-v2 dataset. 


\noindent \textbf{A simple strategy based on gradient norm.}
Motivated by our findings in gradient norm, we consider estimating loss weights based on the gradient's norm, namely \textbf{AvgNorm}. The goal is to scale the norm of each task gradient to a constant or a similar value.

One straghtforward idea is to scale the norm of each task gradient to a constant, \eg, 1. Suppose that there are  \( T \) tasks in total. For the \( i \)-th task (\( i = 1, 2, \dots, T \)), the gradient of the \( i \)-th task loss $\mathcal{L}_i$ with respect to the shared backbone layer is denoted as \( \nabla_{\theta} \mathcal{L}_i \), where \( \theta \) represents the shared parameters.



The loss weight is set according to the reciprocal of the gradient norm of each task: \( w_i = \frac{1}{\|\nabla_{\theta} \mathcal{L}_i\|} \).

However, when the gradient norm of some tasks can be extreme (either close to 0 or very large), \( w_i \) can be highly unstable.
To address the above issues, we instead scale the norm of each gradient to a relative value which changes dynamically during training, \ie, the norm of the aggregated gradients of all tasks \( S = \left\|\sum_{i=1}^{N} \nabla_{\theta} \mathcal{L}_i \right\| \).
And the loss weight is:
\begin{equation}
w_i = \frac{S}{\|\nabla_{\theta} \mathcal{L}_i\|}.
\end{equation}
Here, \( S \) represents the overall strength (norm) of the gradients of all tasks. When the gradient norm \( \|\nabla_{\theta} \mathcal{L}_i\| \) of a certain task is extreme, \( S \) will reflect the ``combined contribution'' of the gradients of the other tasks, thereby suppressing the extreme changes in the weights.

\begin{figure}[!ht]
  \centering
  \includegraphics[width=0.48\textwidth]{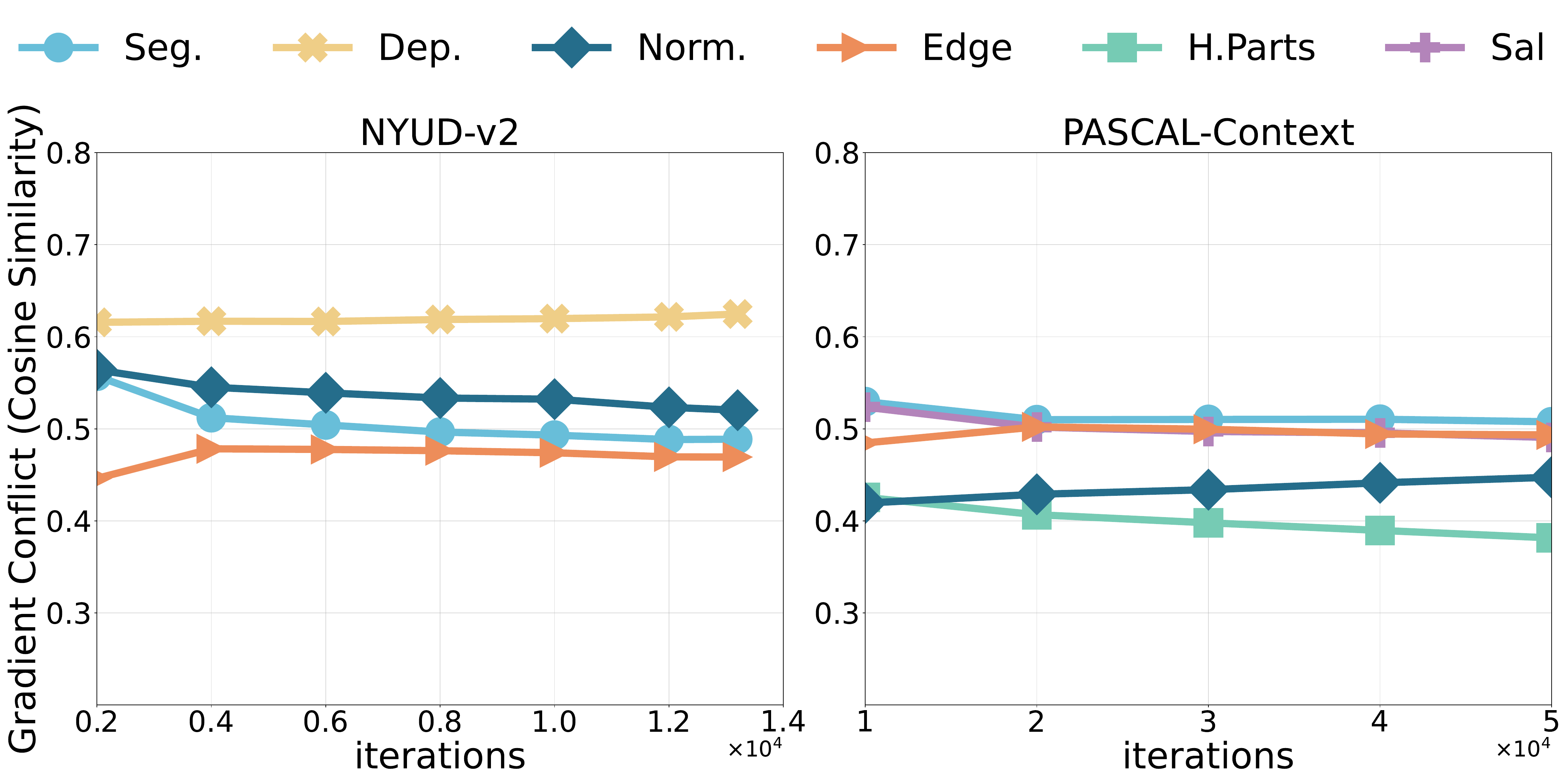}
  \caption{Gradient conflict between the gradient contributed by each individual task and the aggregated gradient for our strategy on NYUD-v2.}
  \label{fig:ours avg grad conflict nyu}
\end{figure}

We compare the AvgNorm to the Grid Search on NYUD-v2 and PASCAL-Context with Vanilla MTL and MTAN and report the results in \cref{tab:ours and grid search on nyud} and \cref{tab:ours and grid search on pascal}. We find that updating the loss weights based on our AvgNorm obtains similar results to the grid search loss weight strategy, \ie a comparable $\Delta_m$ performance is obtained,  with also comparable results on individual tasks (Norm.). 
We also report the average values of loss weights for each task in training of our AvgNorm and we are surprised to find that the loss weights obtained by our method are highly similar to those obtained by grid search, while our method can be performed automatically and much efficient. However, we can still see some different in the loss weights estimated by us. We believe this is due to the fact that the grid search strategy is not really able to find a loss weights that are out of the pre-defined range. In addition, we compute the cosine similarity between the gradient contributed by each individual task and the aggregated gradient in our strategy, as is shown in~\cref{fig:ours avg grad conflict nyu}. We find that the cosine similarity converges to a similar value, which is consistent with previous observations. This proves that our strategy is effective in overcoming the task imbalance problem.

As the grid search strategy is proved to be an effective but cumbersome strategy, the results indicate that our strategy can quickly find a set of relatively beneficial loss weights, overcoming the limitations of the grid search strategy. Secondly, when using MTAN as the backbone, our method outperforms all other methods and achieves the best performance (-0.43\%) on the PASCAL-Context dataset. It has been found experimentally that using the gradient of the last layer of the shared backbone can be sufficient. This again suggests that AvgNorm is simple, effective and computationally efficient.

\begin{table*}[t]
    \centering
    \renewcommand{\arraystretch}{1.2}
    \caption{Performance and loss weight of Grid search and our strategy on NYUD-v2 with ResNet-18. We counted the average of the loss weights obtained by our strategy throughout the training process. $\lambda_t$ is the loss weight of each task. }
    \label{tab:ours and grid search on nyud}
    \scalebox{1}{
    \begin{tabular}{cc|cc|cc|cc|cc|c}
        \toprule
        \multicolumn{1}{c}{\multirow{3}{*}{Backbone}} & \multirow{3}{*}{Method}& \multicolumn{9}{c}{NYUD-v2}\\
        \cmidrule{3-11}&& \multicolumn{2}{c|}{Seg.}& \multicolumn{2}{c|}{Dep.}& \multicolumn{2}{c|}{Norm.}& \multicolumn{2}{c|}{Edge}& \multirow{2}{*}{$\Delta_m$\%↑}\\
         && mIoU↑ &$\lambda_t$& RMSE↓ &$\lambda_t$& mErr↓ &$\lambda_t$& odsF↑  &$\lambda_t$& \\
        \midrule
        \multirow{2}{*}{ResNet-18}
        &Grid Search  & 39.87&1& \textbf{0.5999}&1& 22.74&2& 72.40&50& -0.15\\
 & ours& \textbf{40.04}&1.22& 0.6011&2.47& \textbf{22.43}&3.97& \textbf{72.70}&114.00&\textbf{+0.37}\\
 \midrule
 \multirow{2}{*}{MTAN}

 & Grid Search  & \textbf{40.27}&1& 0.6020&1& \textbf{22.02}&2& 73.70&50&+1.29\\
     & ours& 40.20&1.22& \textbf{0.5949}&2.53& 22.23&3.87& \textbf{73.80}&107.51&\textbf{+1.33}\\
        \bottomrule
    \end{tabular}
    }
\end{table*}

\begin{table*}[t]
    \centering
    \renewcommand{\arraystretch}{1.2}
    \caption{Performance and loss weight of Grid search and our strategy on PASCAL-Context with ResNet-18. }
    \label{tab:ours and grid search on pascal}
    \scalebox{1}{
    \begin{tabular}{cc|cc|cc|cc|cc|cc|c}
        \toprule
        \multicolumn{1}{c}{\multirow{3}{*}{Backbone}} & \multirow{3}{*}{Method}& \multicolumn{11}{c}{PASCAL-Context}\\
        \cmidrule{3-13}&& \multicolumn{2}{c|}{Seg.}& \multicolumn{2}{c|}{H. Parts}& \multicolumn{2}{c|}{Norm.}& \multicolumn{2}{c|}{Sal.}& \multicolumn{2}{c|}{Edge}& \multirow{2}{*}{$\Delta_m$\%↑}\\
         && mIoU↑ &$\lambda_t$& mIoU↑  &$\lambda_t$& mErr↓  &$\lambda_t$& maxF↑  &$\lambda_t$& odsF↑  &$\lambda_t$& \\
        \midrule
        \multirow{2}{*}{ResNet-18}
        &Grid Search  & 65.10 &1& 59.28 &2& \textbf{14.39} &5& 83.91 &5& 65.50 &50& -2.64\\
 & ours& \textbf{65.94}&1.77& \textbf{59.74}&3.57& 14.67&11.42& \textbf{84.38}&3.82& \textbf{66.50} &155.78&\textbf{-2.38}\\
 \midrule
 \multirow{2}{*}{MTAN}

 & Grid Search  & 65.46 &1& 60.47 &2& \textbf{14.14} &5& 84.43 &5& 67.40 &50&-1.22\\
     & ours& \textbf{67.16} &1.8& \textbf{61.24} &2.7& 14.32 &7.3& \textbf{84.54} &2.5& \textbf{68.30} &85.7&\textbf{-0.43}\\
        \bottomrule
    \end{tabular}
    }
\end{table*}

\section{Conclusions}\label{sec:conclude}


In this paper, we presented a systematic study of the task imbalance problem in multi-task dense prediction learning. Our analysis confirms that while a extensive grid search for loss weights is effective, existing optimization (MTO) methods are often inconsistent, and advanced architectures still rely on this costly tuning. Furthermore, our investigation into foundational factors like Vision Foundation Models and data properties revealed they offer limited mitigation. The central insight from our work is the strong correlation between imbalance and the norm of task-specific gradients. Based on this finding, we demonstrate that a simple principle of scaling losses by their gradient norm achieves comparable performance to an exhaustive grid search. Our work suggests the path to robust MTL lies in a principled understanding of gradient dynamics, and we hope our analysis contributes to this deeper understanding.




\bibliographystyle{IEEEtran}
\bibliography{ref}

\clearpage

{\appendices
\section*{Experimental setups}
In this section, we provide more details about our experimental implementation in addition to those discussed in the paper.
\subsection*{Datasets and data augmentation} 
We conduct the experiments on three multi-task datasets: PASCAL-Context\cite{pascal}, NYUD-v2\cite{nyud}, and Replica\cite{replica}. We take the same data augmentation as~\cite{atrc,invpt}. We use random scaling with factor between 0.5 and 2.0, random horizontal flipping, and random color jittering. Image normalization is also applied throughout both the training and evaluation phases. For NYUD-v2, All images are resized to 448 × 576 resolution as in~\cite{invpt}(except for DINO-v2, we cropped the image in DINO-v2 to 488 × 574 pixels to fit the input requirements of the model). For PASCAL-Context and Replica, we resize the images to 512 x 512 pixels (except for DINO-v2, we cropped the image in DINO-v2 to 518 × 518 pixels). 
\subsection*{Loss functions and task weight} Following previous work~\cite{Maninis_2019_CVPR, invpt}, we employ task-specific loss functions. For semantic segmentation and human parsing, we use the cross-entropy loss, while for saliency detection, we apply the balanced cross-entropy loss. Surface normal estimation and monocular depth estimation are optimized using the balanced L1 loss. For edge detection, we use the weighted binary cross-entropy loss, with a weight of 0.95 to positive pixels and 0.05 to negative ones. We compute a weighted sum of the task-specific losses. For Grad Search stratigy, we refer to previous studies~\cite{simon_survey,invpt} and re-search the loss weight space of the task. For NYUD-v2 dataset, the loss weights for Seg., Dep., Norm., and Edge tasks are set to 1, 1, 2, and 50, respectively. For PASCAL-Context, Seg., H.Parts, Sal., Norm., and Edge tasks are set to 1, 2, 5, 5, and 50, respectively.
\subsection*{Training details}We follow the implementation in prior work~\cite{simon_survey,invpt}. For ResNet, we train the model on NYUD-v2 and PASCAL-Context for 100 and 60 epochs, respectively. We evaluate after each training epoch and take the best model to report the results. For HRNet and ViT, we train the model on NYUD-v2 and PASCAL-Context for 40k iterations. We evaluate after every 1k iterations and take the best model to report the results. On the Replica, we use InvPT as the base model and follow all the settings of the ViT backbone. We choose the best model on the validation set and test it on the test set.

\subsection*{Others settings}
For the different VFMs as model initialization, we uniformly used ViT-B/16 backbone, we called the relevant model weights from the Timm library as shown in \cref{tab: VFMs details}. In order to best present the effectiveness of all MTOs, we searched for hyperparameters of MTOs on different datasets.
For GradNorm, we search through $\alpha \in[0.1, 0.2, 0.5, 1.0, 1.5, 2.0]$. For DWA, we search through $ T\in[0.5, 1.0, 1.5, 2.0, 2.5, 3.0, 4.0]$. For other MTOs, we use the default settings for their deployment.

\begin{table}
    \vspace{-2cm}
    \centering
    \renewcommand{\arraystretch}{1.5}
    \caption{Model zoo of VFMs.}
    \label{tab: VFMs details}
    \scalebox{1}{
    \begin{tabular}{c|c}
        \toprule
         Model & Name\\
        \midrule
        \multirow{3}{*}{ViT\cite{vit}}&  vit\_base\_patch16\_384.augreg\_in1k\\
 &vit\_base\_patch16\_224.augreg\_in21k\\
 &vit\_base\_patch16\_384.augreg\_in21k\_ft\_in1k\\
\hdashline
 CLIP\cite{clip}&vit\_base\_patch16\_clip\_384\\
 SAM\cite{sam}&samvit\_base\_patch16\\
 DINO\cite{DINO}&vit\_base\_patch16\_224.dino\\
        DINO-v2\cite{dinov2}&  vit\_base\_patch14\_reg4\_dinov2\\
        \bottomrule
    \end{tabular}
    }
\end{table}

\begin{table}
    \vspace{-5cm}
    \centering
    \renewcommand{\arraystretch}{1.5}
    \caption{Training hyperparameters setup.}
    \label{tab: training details}
    \scalebox{1}{
    \begin{tabular}{c|ccc}
        \toprule
         Hyperparameters& ResNet& HRNet& ViT\\
        \midrule
        \#GPUs& 1& 1& 2\\
  Total batch size& 6& 6& 4\\

   Optimizer& Adam& Adam& AdamW\\
        Base learning rate& 1e-4& 1e-4& 2e-5\\
        Weight Decay& 1e-4& 1e-4& 1e-6\\
        LR scheduler& Poly& Poly& Poly\\
        Steps& 100, 60 epochs& 40k iterations& 40k iterations\\
 Gradient clipping& 0& 0&10\\
        Test round& 1 epoch& 1k iterations& 1k iterations\\
        \bottomrule
    \end{tabular}
    }
    \vspace{10cm}
\end{table}


\clearpage
\section*{More Results}
In this section, we provide more detailed results including gradient conflicts, gradient norms, and more.
We count the values in each training iteration. For the presentation of the results, we take the average values for visualization (2k iterations and 10k iterations for NYUD-v2 and PASCAL-Context dataset, respectively).

\subsection{Gradient conflict}
\subsubsection*{Gradient similarity between tasks}
For the similarity of the gradient between tasks, we consider two ways to calculate the similarity of the gradient. The first is similar to~\cite{pcgrad}, we use the cosine similarity to measure the conflict between different task gradients. The second is similar to~\cite{cagrad}, we use the dot product to measure the conflict between different task gradients.

\textbf{1) Cosine similarity:} It only takes into account the angular relationship between the gradients. We report the results of the Uniform, Grad Search, CAGrad, and NashMTL on NYUD-v2 and PASCAL-Context datasets, as shown in \cref{fig:nyud grad conflict of two tasks} and \cref{fig:pascal grad conflict of two tasks}. 

\textbf{Finding:} As training progresses, most task pairs are convergent orthogonal. In addition, there seems to be a strong correlation between gradient similarity and task type, although all tasks were shown to be similar tasks that can be trained together. For example, the similarity between segmentation type tasks in PASCAL-Context (e.g., Seg., H. Parts, and Edge) maintained high values during training. This stems from the fact that they are more similar in terms of task type. However, this observation does not seem to be clearly related to optimization imbalance in MTL.

\textbf{2) Dot product:} It takes into account both the angle and the size of the gradient. We report the results of Uniform, Grad Search, CAGrad, and NashMTL on NYUD-v2 and PASCAL-Context datasets, as shown in \cref{fig:nyud dot} and \cref{fig:pascal dot}.

\textbf{Finding:} We find that the gradients of the different tasks tend to be orthogonal, an observation that is the same as the result for cosine similarity. However, the difference is that in the NYUD-v2 dataset, half of the task pairs in the Uniform method are already orthogonal at the beginning of training and remain orthogonal throughout the training process, e.g., Seg. and Edge. This suggests that although Seg. and Edge are very similar in terms of direction (gradient angle), the optimization of the Edge task, due to the effect of the gradient size is poorer, which leads to imbalance. The optimal loss weights obtained from the grid search effectively avoid negative optimization of the Edge task. And the results of CAGrad and NashMTL show aggregation. The Grid Search, CAGrad and NashMTL all achieved relatively better results, but Grid Search presents a different phenomenon than these two MTOs, and thus we did not observe a valid correlation between dot product and imbalance between tasks.

Similarly, since it is more difficult to optimize on the PASACL-Context dataset (more tasks), the vast majority of task pairs in the Uniform are already orthogonal at the start of training (except for Seg. and H.Parts). We have similarly observed the contribution of Grid Search in this regard. In addition we likewise observe a similar phenomenon for Grid Search, CAGrad and NashMTL.

\subsubsection*{Gradient similarity between each task and aggregated direction}

For the cosine similarity between the gradients of each task and the aggregated gradients, we complemented the results with a visualization of the PASCAL-Context, see \cref{fig:avg_grad_conflict_pascal}.

\textbf{Finding:} Optimization on the Pascal-Context dataset is more challenging as it contains more tasks. We find that the similarity of the tasks in the Grid Search converges again to the same value except for the Norm. task. Except for the sal. task, CAGrad is more clustered for other tasks. For NashMTL this convergence is even more pronounced.

\textbf{Conclution:} This observation suggests that, while MTL methods may encourage orthogonal gradients, this alone does not necessarily resolve task unbalance. And our analysis suggests that the key to overcoming task unbalance lies in the similarity between each task's gradient and the aggregated gradient. A more balanced contribution from each task, as indicated by a cosine similarity closer to an identical value, may be indicative of a more effective multi-task learning approach.

\subsection{Gradient norm}

In \cref{fig:grad_norm_pascal}, we additionally illustrate the change in gradient norm for each task during the training process on the PASCAL-Context dataset. We also observed the results for Single-task, as shown in \cref{fig:grad_norm_stl}

\textbf{Finding:} In the PASCAL-Context dataset, except for the Norm. task, the vast majority of tasks focus the gradient L2 criterion on a similar value through a grid search strategy. Since the PASCAL-Context dataset is more challenging, we believe that it is possible that our grid search loss weighting strategy did not reach the Pareto frontier. We observe that CAGrad and NashMTL converge the gradient norms of all tasks to a similar value. Furthermore, we do not observe a potential correlation between Single-task and multi-task in terms of gradient norm.

\textbf{Conclution:} Limiting the gradient norms across tasks to a relatively consistent range can alleviate the problem of task imbalance. CAGrad and NashMTL balance the gradient sizes well, but our method does not require modification of the gradient information and is computationally less expensive while achieving comparable performance to existing methods.

\begin{figure*}[!ht]
  \centering
  \includegraphics[width=1\textwidth]{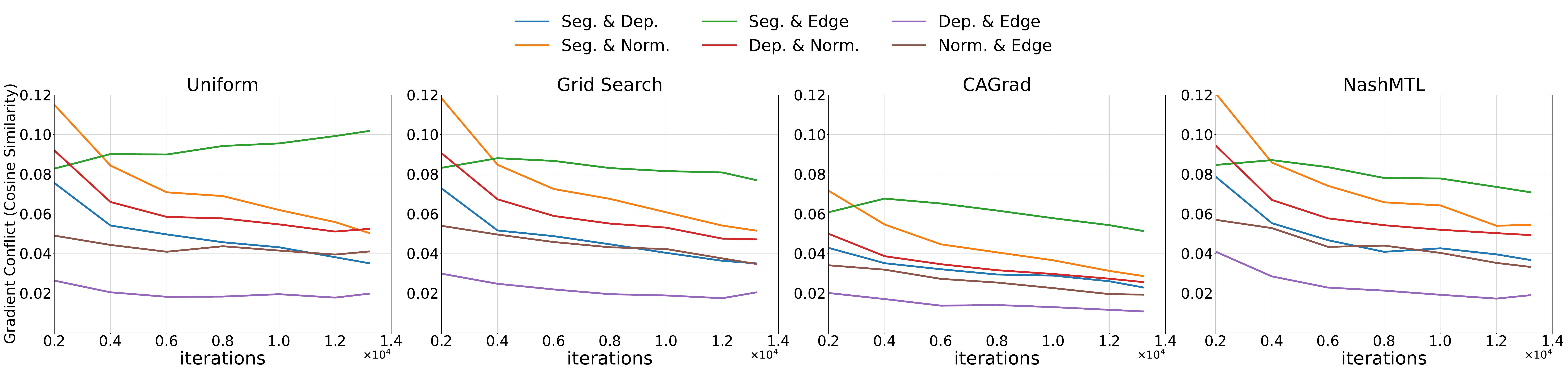}
  \caption{Gradient conflict between two tasks for Uniform, Grid Search, CAGrad and NashMTL during the training process on NYUD-v2.}
  \label{fig:nyud grad conflict of two tasks}
\end{figure*}

\begin{figure*}[!ht]
  \centering
  \includegraphics[width=1\textwidth]{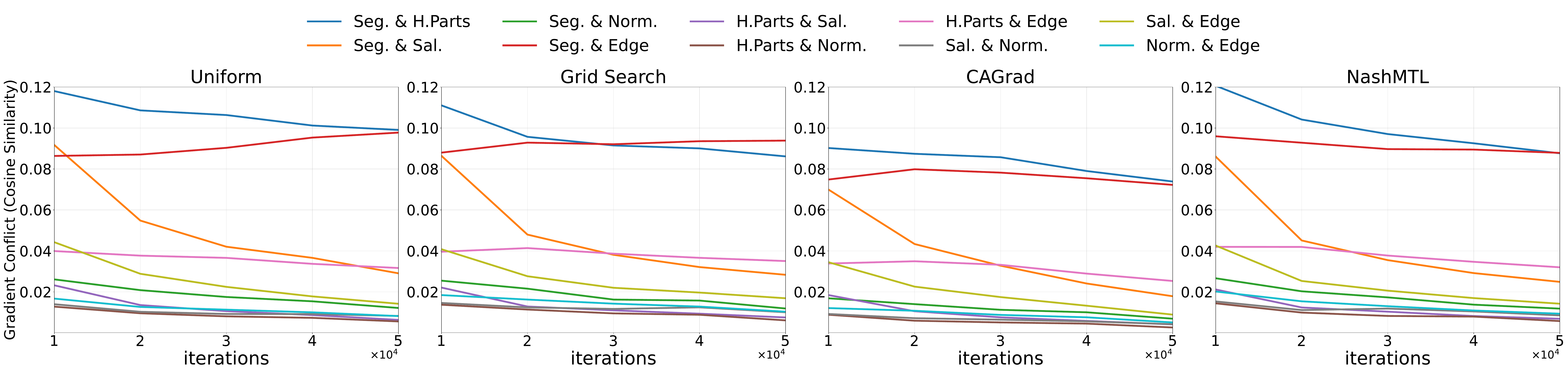}
  \caption{Gradient conflict between two tasks for Uniform, Grid Search, CAGrad and NashMTL during the training process on PASCAL-Context.}
  \label{fig:pascal grad conflict of two tasks}
\end{figure*}

\begin{figure*}[!ht]
  \centering
  \includegraphics[width=1\textwidth]{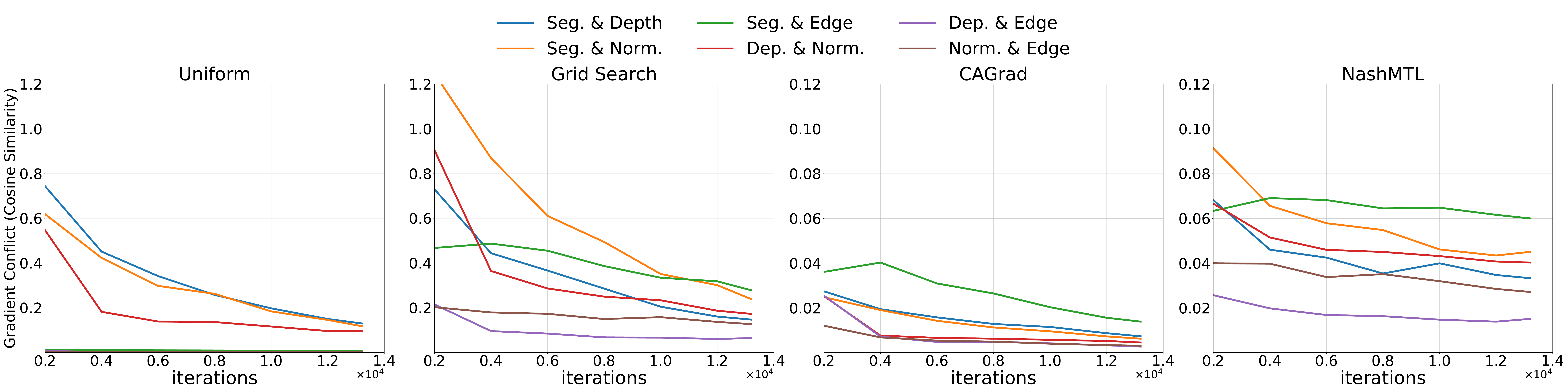}
  \caption{Dot product between two tasks for Uniform, Grid Search, CAGrad and NashMTL during the training process on NYUD-v2.}
  \label{fig:nyud dot}
\end{figure*}

\begin{figure*}[!ht]
  \centering
  \includegraphics[width=1\textwidth]{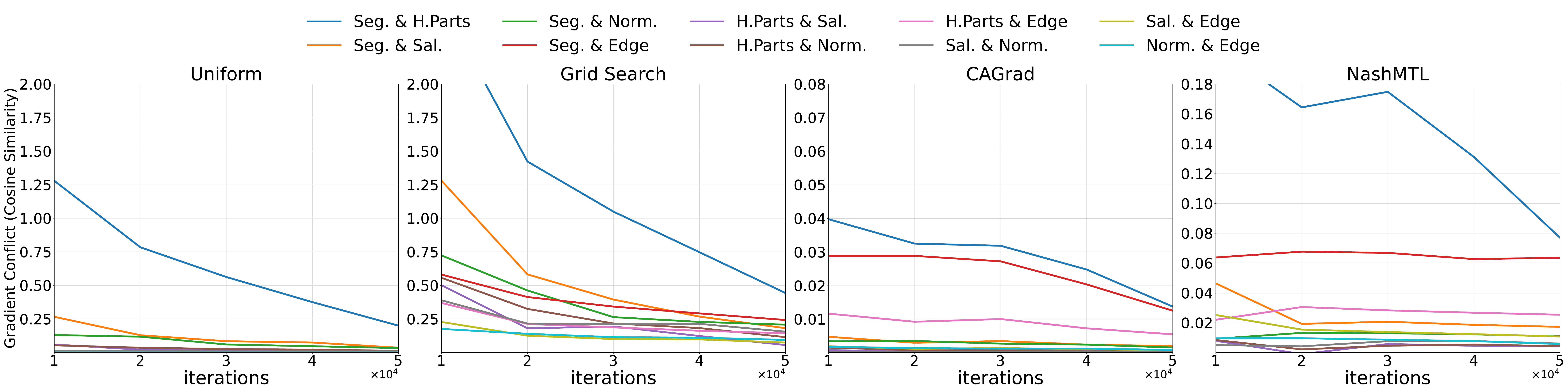}
  \caption{Dot product between two tasks for Uniform, Grid Search, CAGrad and NashMTL during the training process on PASCAL-Context.}
  \label{fig:pascal dot}
\end{figure*}

\begin{figure*}
  \centering
  \includegraphics[width=1\textwidth]{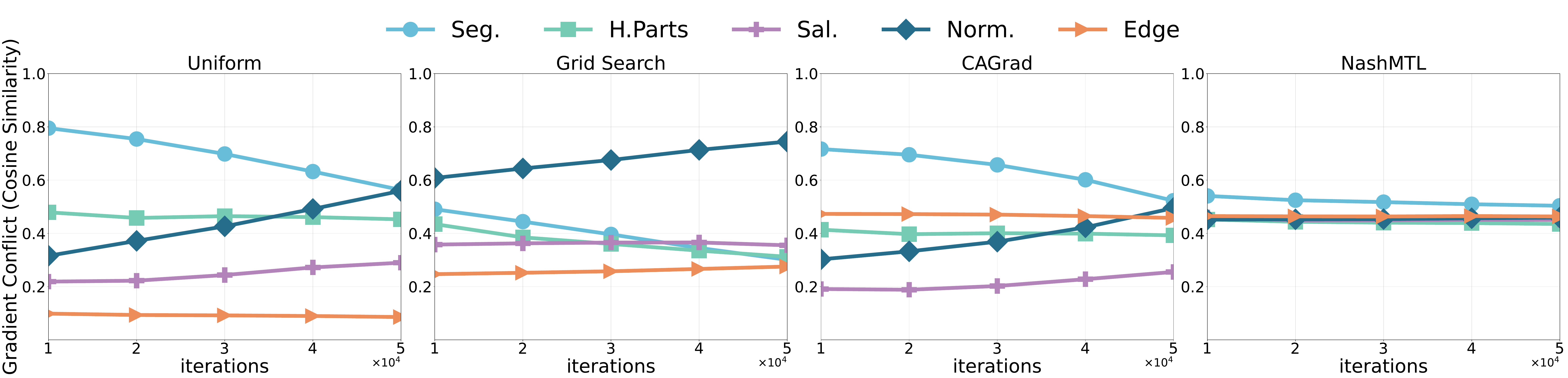}
  \caption{Gradient conflict between the gradient contributed by each individual task and the aggregated gradient for Uniform, Grid Search, CAGrad and NashMTL on PASCAL-Context.}
  \label{fig:avg_grad_conflict_pascal}
\end{figure*}

\begin{figure*}
  \centering
  \includegraphics[width=1\textwidth]{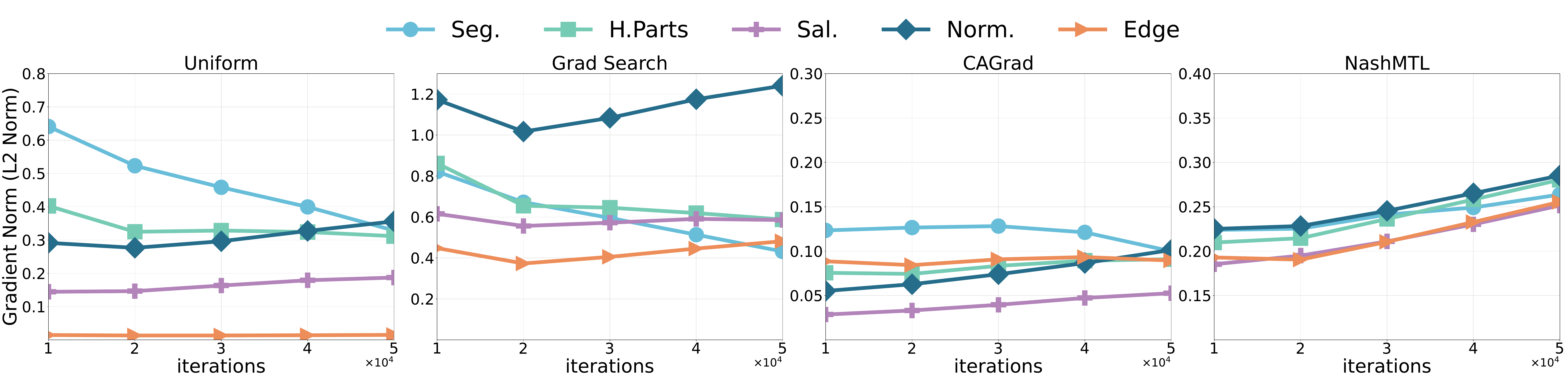}
  \caption{Gradient norm (last convolutional layer of the backbone) of different methods for each task on PASCAL-Context.}
  \label{fig:grad_norm_pascal}
\end{figure*}

\begin{figure*}
  \centering
  \includegraphics[width=0.7\textwidth]{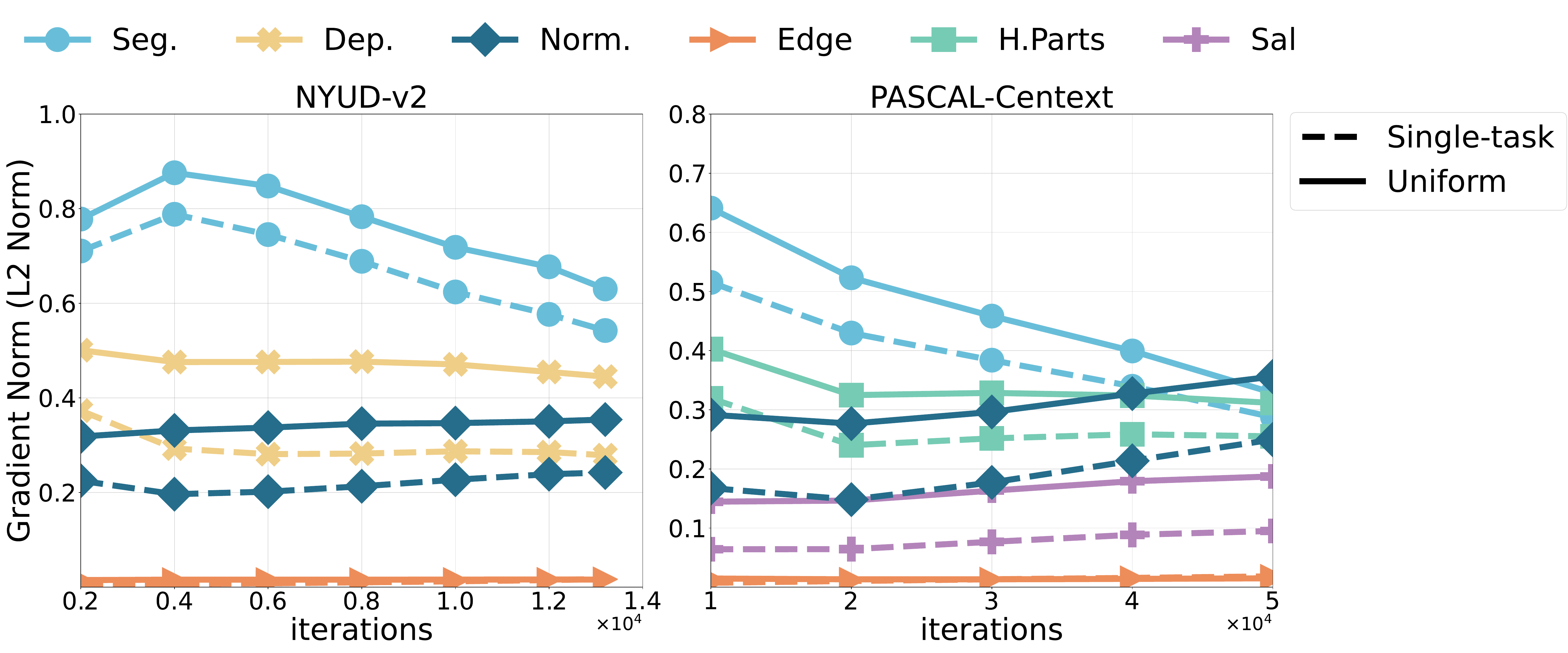}
  \caption{Gradient (w.r.t. the last layer of the encoder) norm for each task in Uniform, Grid Search, CAGrad, and NashMTL on PASCAL-Context.}
  \label{fig:grad_norm_stl}
\end{figure*}

\end{document}